# From Canopy to Collision: A Hybrid Predictive Framework for Identifying Risk Factors in Tree-Involved Traffic Crashes

**Abdul Azim (Corresponding Author)**
Department of Civil Engineering
Rajshahi University of Engineering & Technology
Rajshahi-6204, Bangladesh
Email: azimsarkar03@gmail.com
ORCID: 0009-0003-5784-6488

**Ahmed Hossain, Ph.D.**
Traffic Safety Analysis Engineer, Traffic Safety Group
Multimodal Planning Division (MPD)
Arizona Department of Transportation (ADOT)
1615 W. Jackson St
Phoenix, AZ 85007
Email: ahmedhossain09@outlook.com
ORCID: 0000-0003-1566-3993

**Soumyadip Maitra**
Department of Civil Engineering
Rajshahi University of Engineering & Technology
Rajshahi-6204, Bangladesh
Email: soumyadipmaitra50@gmail.com
ORCID: 0009-0001-3785-3627

**Panick Kalambay, Ph.D.**
Assistant Professor
Department of Transportation Studies
College of Science, Engineering and Technology
Texas Southern University
3100 Cleburne Street, Houston, TX, 77004
Email: panick.kalambay@tsu.edu

## ABSTRACT

Tree-involved crashes represent a critical subset of run-off-road (ROR) collisions, often resulting in fatal or severe injuries due to high-energy impacts. This study develops a comprehensive analytical framework to identify and quantify risk factors contributing to crash severity in tree-involved collisions using the Crash Report Sampling System (CRSS) database spanning 2020-2023. The modeling framework follows a multi-step process. First, a machine learning based classification model (CatBoost) identifies key factors associated with binary crash injury severity (KA: fatal or incapacitating injury versus BC: non-incapacitating or possible injury). Second, SHapley Additive exPlanations (SHAP) tool is used to quantify and visualize the marginal effects of top influential factors on crash severity. Third, a binary logistic regression model estimates factor effects and validates SHAP-derived importance measures. Finally, SHAP interaction plots examine the combined effects of key contributing factors. Results reveal restraint non-use as the most influential predictor, with unrestrained occupants nearly three times more likely to experience severe outcomes due to ejection risk. Vehicle age, speeding violations, and driver impairment demonstrate substantial effects, reflecting reduced crashworthiness, increased impact forces, and reduced control capabilities. Critical interactions emerge between lighting conditions and vehicle age, speeding and lighting conditions, restraint use and vehicle age, and road surface and speeding, demonstrating additive risk effects with specific interactions. These findings provide critical insights for targeted safe system-based interventions, including enhanced seat belt enforcement, speed management in reduced visibility conditions, and vehicle fleet modernization.

## 1. INTRODUCTION

Tree-involved crashes represent a critical and persistent hazard in roadway safety, characterized by disproportionately severe injury and fatality outcomes compared to other crash types. National-level evidence from the Fatality Analysis Reporting System (FARS) indicates that tree-involved crashes represent the most harmful outcome in roadway departure crash events. Analysis of FARS data from 2016 – 2018 shows that collisions with trees accounted for 10,697 roadway-departure fatalities during this period, underscoring their disproportionate contribution to traffic-related mortality [1]. A more recent review of U.S. traffic fatality data from 2020 to 2023 reveals that collisions with trees persistently account for approximately one-quarter of all fatalities involving fixed objects (more details in Figure 1), underscoring the ongoing and specific danger posed by roadside trees in roadway safety. These statistics underscore the critical role of tree-involved crashes in roadway departure fatalities and highlight the urgent need for focused investigation into the crash contributing factors that contribute to their severity.

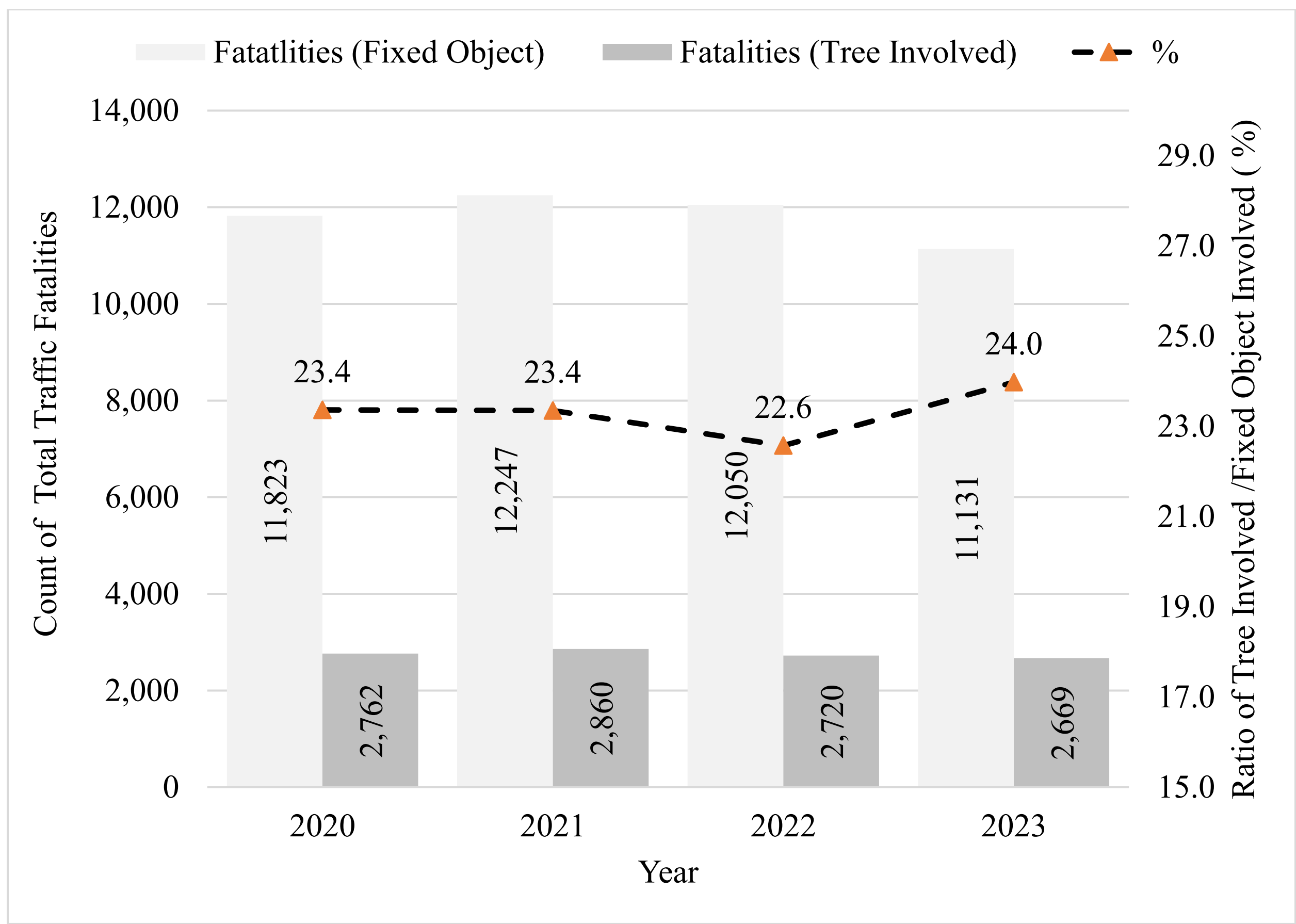

**Figure 1. Fixed Object and Tree Involved Fatalities in the US (2019-2023)**

The high severity of tree-vehicle crashes results from the fundamental biomechanical incompatibility between rigid tree structures and vehicle energy-absorbing systems [2]. Trees function as narrow, non-deformable objects that concentrate collision forces onto a localized area of the vehicle chassis, bypassing the main structural load paths designed for distributed impacts [3]. Unlike forgiving roadside infrastructure such as breakaway utility poles or energy-absorbing guardrails, which are designed to dissipate kinetic energy, trees possess negligible deformation capacity [4]. This rigidity causes the vehicle to deform and crush, often resulting in serious intrusion into the occupant compartment. When the survival space is reduced in this way, injuries, especially to the chest and legs, are more severe and are associated with higher injury severity than in crashes with deformable barriers [5].

The risk of tree-vehicle collisions arises from a complex interaction of vehicle characteristics, roadway geometry, environmental conditions, and driver behavior [6,7]. This multifactorial nature requires comprehensive analytical methods capable of identifying non-linear relationships and interaction effects among contributing factors. Although tree-involved crashes represent a critical subset of run-off-road (ROR) and fixed-object collisions, they have received limited attention in crash severity research compared to the broader fixed-object literature. In addition, relatively few studies have focused explicitly on advanced analytical approaches to capture their complex, multifactorial nature.

To address this gap, this study investigates the contributing factors of tree-involved crashes using a hybrid methodological framework. By integrating CatBoost, a gradient-boosting decision tree algorithm, with SHapley Additive exPlanations (SHAP) and binomial logistic regression, this research aims to leverage the predictive power of machine learning (ML) algorithm while retaining the interpretability of classic regression model required for a deeper understanding of the contextual factors. Overall, the objective of this study is to (a) identify and quantify the key determinants of crash injury severity in tree-involved collisions, and (b) investigate non-linear relationships and interaction effects among significant variables. The findings of this research are expected to provide a more comprehensive understanding of the core factors associated with tree-involved crashes and to support informed decision-making in developing problem-specific safety countermeasures.

## 2. LITERATURE REVIEW

Several previous studies have highlighted tree-involved crashes as a persistent safety concern, focusing on various contextual factors within the broader context of fixed-object collision investigations. For instance, a study conducted in Washington State found that collisions with trees or poles significantly increased the likelihood of fatal injuries [7]. Another study from the same state reported that a collision with trees elevates the probability of a passenger's fatality by about 250% in fixed-object crashes [8]. Vehicle type also plays a dominant role in tree-involved collisions. For example, a previous study on motorcycle crashes reported that collisions with roadside trees were associated with substantially higher fatality risk than crashes involving only ground impact [9]. Using FARS and General Estimates System (GES) databases, this study found that motorcycle collisions with trees were nearly 15 times more likely to be fatal than collisions with the ground. Similarly, a nationwide analysis of single-vehicle motorcycle crashes in the United States reported that fixed-object collisions contribute substantially to motorcyclist fatalities, with tree impacts associated with approximately 3.5 times higher fatality risk than barrier collisions [10]. In addition, a recent study of fixed-object crashes in Texas found that injury severity varies considerably across fixed-object types, identifying trees and shrubs as key contributors to higher injury risk [11]. A temporal analysis of fixed-object passenger-car crashes in Oregon found that collisions with trees consistently increased injury severity across the pre, during, and post-pandemic periods, with severe injury emerging as the most likely outcome in tree-involved crashes [12].

Tree-involved crashes stem from the combined effects of vehicle, driver, roadway, and environmental factors, as shown in prior research. Older vehicles substantially increase the risk of severe injury compared to newer ones, due to weaker structural design and outdated safety features [2]. Driver demographic characteristics also play a significant role. Younger male drivers are disproportionately more involved in tree collisions, and higher travel speeds, including both exceeding posted speed limits and driving too fast for prevailing conditions, are strongly associated with increased injury severity [13,14]. The driver's physical condition plays a vital role in tree-involved crashes. For example, alcohol and drug involvement substantially raises the risk of fatalities in tree-involved crashes, with alcohol-involved collisions showing significantly higher fatality rates than sober-driver crashes [6]. Roadway geometry and design strongly influence tree-involved crash outcomes. Tree-involved collisions occur most often on two-lane undivided roads and horizontal curves, with most struck trees within approximately 30 feet of the roadway, and inadequate clear zones consistently linked to fatal crashes [13,15]. Rural areas experience higher crash frequency and severity than urban areas, and darkness significantly increases the risk of serious injury due to reduced visibility and slow reaction time, while fog and crosswinds heighten crash likelihood by affecting vehicle control and driver judgment [16,17]. Together, these studies show that tree-involved crash severity results from the combined influence of impact speed, vehicle protection, tree size, roadway design, visibility, and driver behavior, requiring analytical methods capable of identifying nonlinear relationships and interactions among these factors.

Previous studies have widely explored methodologies for predicting crash severity in tree-involved crash investigations, employing a spectrum of models. Table 1 summarizes key details in previous tree-involved crash investigations.

**Table 1. Previous Tree-Involved Crash Investigations**

| Study | Year | Location | Specific Focus | Model(s) Used |
|---|---|---|---|---|
| [6] | 1990 | United States (urban) | Trees (urban context) | Descriptive/field investigation; frequency & pattern analysis |
| [18] | 1999 | United States | Trees, utility poles, guardrails (side) | NASS/FARS data analysis; velocity change estimation |
| [19] | 2000 | United States | Trees and poles (side impact) | Side impact risk method; Head Injury Criterion (HIC) |
| [20] | 2008 | Europe (multi-country) | Trees and rigid roadside objects | MADYMO simulation; Finite Element Analysis (FEA) |
| [13] | 2009 | United States (SC) | Trees | Qualitative and quantitative factor analysis; Roadside Safety Analysis Program (RSAP) |
| [21] | 2013 | United States (MA) | Trees and fixed roadside objects (ROR) | Crash severity analysis (statistical/regression-based) |
| [22] | 2017 | United States | Trees and roadside vegetation | Crash frequency and severity modeling (literature synthesis) |
| [23] | 2020 | China | Trees (highway) | PC-Crash Simulation; Chest Resultant Acceleration (CRA); Acceleration Severity Index (ASI) |
| [2] | 2020 | Czech Republic | Trees (single vehicle) | Multinomial Logistic Regression; Generalized Linear Mixed Models (GLMM) |

*Note: NASS = National Accident Sampling System; HIC = Head Injury Criterion; CRA = Chest Resultant Acceleration; ASI = Acceleration Severity Index; RSAP = Roadside Safety Analysis Program.*

From a general perspective of crash severity investigations, statistical models, such as logistic regression, negative binomial regression, ordered probit, and multinomial logit models, have been widely employed to identify significant contributing factors and quantify their effects on crash outcomes [24–28]. Random parameter logit models are also widely used in this field due to their interpretability and ability to quantify the effects of explanatory variables while accounting for unobserved heterogeneity [29–31]. In recent years, the adoption of ML techniques in crash modeling has also surged, primarily due to their enhanced predictive capabilities and ability to capture nonlinear relationships and complex interactions [32]. Commonly applied ML methods in crash severity analysis include Random Forest (RF), Support Vector Machine (SVM), gradient boosting models, and neural networks [33–36]. Recent injury severity studies increasingly employed CatBoost-based gradient boosting models because of their effective handling of categorical variables and high predictive accuracy with limited parameter tuning [37–39]. Comparative analyses have shown that while statistical models provide critical insights into crash contributing factors, ML models excel at detecting intricate crash patterns and uncovering hidden correlations among variables [40].

Despite their individual strengths, both approaches have limitations when applied in isolation. Statistical models often struggle with higher-order interactions, whereas ML models typically lack inferential interpretability due to their black box nature. Recognizing this, a growing body of research has adopted hybrid modeling frameworks that integrate statistical and ML techniques to leverage the complementary strengths of both approaches. In these frameworks, ML methods are used for feature selection and to model nonlinear effects, while statistical models provide inferential insights and quantify effect sizes [41–46]. This integration improves predictive accuracy while retaining the interpretability needed for safety analysis and policy applications.

### 2.1 Research Gap and Study Objectives

Based on an extensive review of the literature on tree-involved crash investigations, it is evident that a comprehensive exploration of the core crash-contributing factors using a hybrid analytical approach is not adequately addressed. Most existing studies rely on traditional statistical methods which often fail to capture complex nonlinear relationships and interaction effects among roadway, vehicle, driver, and environmental factors. Moreover, limited attention has been given to integrating data-driven pattern discovery with rigorous inferential modeling to improve interpretability and practical applicability. As a result, the underlying mechanisms that jointly influence the occurrence and severity of tree-involved crashes remain insufficiently understood, highlighting the need for a unified hybrid framework that combines the strengths of both ML and statistical modeling approaches. The objective of this research is to explore factors contributing to crash severity in tree-involved crashes and to provide context-specific safety recommendations based on the model results.

## 3. METHODOLOGY

### 3.1 Modeling Framework

The proposed modeling framework follows a multi-step analytical process. First, a CatBoost classification model is employed to identify the key factors associated with binary crash injury severity (KA versus BC) in tree-involved collisions. Second, the SHAP tool is utilized to quantify and visualize the marginal effects of the most influential factors on crash severity. Third, guided by the SHAP-based insights, a binary logistic regression model is developed to estimate the effects of selected factors and to compare inferential results with the SHAP-derived importance measures. Finally, SHAP-based interaction plots are generated to examine the combined and interaction effects of the key contributing factors on crash severity.

#### 3.1.1 *CatBoost*

Categorical Boosting (CatBoost) is an open-source gradient boosting algorithm that has demonstrated better predictive performance compared to other publicly available boosting implementations across diverse benchmark datasets [47]. The algorithm was specifically designed to address fundamental limitations in conventional gradient boosting frameworks, particularly in handling categorical variables and preventing target leakage during model training.

The algorithmic foundation of CatBoost introduces two key approaches, ordered boosting and ordered target statistics [48]. Both approaches were developed to address prediction shift, a phenomenon arising from target leakage that affects traditional gradient boosting implementations. The CatBoost model constructs an ensemble of symmetric decision trees iteratively, where the final prediction can be expressed using **Equation 1**:

$$F(x) = \sum_{t=1}^{T} f_t(x, \theta_t) \quad [1]$$

Where $F(x)$ represents the aggregate prediction for input feature vector $x$, $T$ denotes the total number of decision trees in the ensemble, and $\theta_t$ contains the learned parameters for the $t$-th tree.

CatBoost's approach of ordered target statistics ensures that each categorical feature is transformed using only information from preceding observations in a random permutation, thereby preventing target leakage and maintaining model integrity. The ordered boosting mechanism further reinforces this principle by computing gradient estimates exclusively from historical observations, effectively eliminating the conditional shift that compromises traditional gradient boosting implementations [48]. Mathematically, for a categorical feature $k$, the ordered target statistic is computed using a permutation-based approach as shown in **Equation 2**:

$$\hat{x}_i^k = \frac{\sum_{j=1}^{i-1} [x_j^k = x_i^k] \cdot y_j + a \cdot P}{\sum_{j=1}^{i-1} [x_j^k = x_i^k] + a} \quad [2]$$

Where $\hat{x}_i^k$ is the encoded value for categorical feature $k$ of observation $i$, $[\cdot]$ denotes Iverson brackets, i.e., $[x_j^k = x_i^k]$ equals 1 if $x_j^k = x_i^k$ and 0 otherwise, $y_j$ represents the target value of preceding observations, $P$ is the prior value (typically the global target mean), and $a$ is a smoothing parameter controlling the balance between category-specific statistics and the prior.

Given CatBoost's native capability to handle categorical variables prevalent in traffic crash datasets, including road characteristics, environmental conditions, and driver attributes, the algorithm was employed in this study to identify the most contributing factors associated with tree-involved crashes.

### 3.1.2 *SHAP-Based Model Interpretation*

Despite CatBoost's strong predictive capabilities, ensemble machine learning models inherently lack transparency in explaining how individual features contribute to predictions. To overcome this interpretability limitation, we employed SHAP analysis in this study. SHAP is a game-theoretic approach that assigns each feature an importance value representing its contribution to the model's prediction for each observation [49]. The SHAP value for a feature $i$ is computed based on Shapley values which are shown in **Equation 3**:

$$\phi_i = \sum_{S \subseteq N \setminus \{i\}} \frac{|S|!(|N|-|S|-1)!}{|N|!} [f(S \cup \{i\}) - f(S)] \quad [3]$$

Where $\phi_i$ represents the SHAP value for feature $i$, $N$ is the set of all features, $S$ is a subset of features excluding $i$, $f(S \cup \{i\})$ is the model prediction with feature $i$ included, and $f(S)$ is the prediction without feature $i$.

The SHAP summary plot visualizes the overall importance and directional effects of the top 20 features based on model predictions. This plot displays features ranked by their mean absolute SHAP values, with individual points representing observations and colors indicating feature values. The summary plot enables identification of the most influential factors and reveals whether high or low values of each feature increase or decrease the likelihood of crash severity in tree-involved collisions.

Furthermore, SHAP interaction plots were constructed to examine the combined effects of the top interacting feature pairs on crash predictions [50]. SHAP interaction values decompose the total SHAP value into main effects and pairwise interaction effects, computed as shown in **Equation 4**:

$$\phi_{i,j} = \sum_{S \subseteq N \setminus \{i,j\}} \frac{|S|!(|N|-|S|-2)!}{2(|N|-1)!} \nabla_{i,j}(S) \quad [4]$$

Where $\phi_{i,j}$ represents the interaction effect between features $i$ and $j$, and $\nabla_{i,j}(S) = f(S \cup \{i,j\}) - f(S \cup \{i\}) - f(S \cup \{j\}) + f(S)$ captures the discrete second-order difference of the model output when both features are added.

The interaction plots reveal how the effect of one variable on crash occurrence changes depending on the value of another variable, thereby uncovering synergistic or antagonistic relationships between contributing factors. This comprehensive SHAP analysis framework enables a nuanced understanding of the complex factor interactions underlying tree-involved crashes.

### 3.1.3 *Binary Logistic Regression Model*

Building upon the SHAP importance rankings, a binary logistic regression model is developed to statistically quantify the effects of key contributing factors on crash severity outcomes. Binary logistic regression is a statistical technique widely used in traffic safety research to model the relationship between a binary dependent variable (e.g., crash severity, KA vs BC) and multiple independent variables [51]. Prior to model development, multicollinearity among the top-ranked predictor variables was assessed using Cramér's V statistic. Cramér's V is a measure of association between categorical variables, ranging from 0 (no association) to 1 (perfect association), calculated as shown in **Equation 5**:

$$V = \sqrt{\frac{\chi^2}{n \times (k-1)}} \quad [5]$$

Where $\chi^2$ is the chi-square statistic, $n$ is the total number of observations (i.e., crash count), and $k$ is the minimum of the number of rows or columns in the contingency table. Variable pairs exhibiting high correlation were identified, and one variable from each highly correlated pair was removed based on theoretical relevance to tree-involved crash occurrence. This procedure ensures that the remaining predictor variables contribute unique information to the model, thereby avoiding multicollinearity issues that could inflate standard errors and produce unstable coefficient estimates [52].

Following the multicollinearity assessment, the binary logistic regression model is estimated using the remaining predictor variables. The model estimates the probability of an outcome (fatal-severe crash injury) occurring based on the logistic function, expressed as shown in **Equation 6**:

$$P(Y = 1 \mid X) = \frac{1}{1 + \exp\left(-\left(\beta_0 + \sum_{i=1}^{n} \beta_i X_i\right)\right)} \quad [6]$$

Where $P(Y = 1 \mid X)$ represents the probability of a fatal-severe crash occurring, $\beta_0$ is the intercept term, and $\beta_i$ represents the regression coefficient for the $i$-th predictor variable $X_i$. The model can also be expressed in terms of the log-odds (logit) transformation, as shown in **Equation 7**:

$$\ln\left(\frac{P}{1-P}\right) = \beta_0 + \sum_{i=1}^{n} \beta_i X_i \quad [7]$$

Where $\ln(\frac{P}{1-P})$ represents the natural logarithm of the odds of a fatal-severe crash occurring. The model parameters were estimated using maximum likelihood estimation (MLE), which identifies the coefficient values that maximize the likelihood of observing the given data. To interpret the effect of each predictor variable on crash severity, the Odds Ratio (OR) is computed by exponentiating the regression coefficients, as shown in **Equation 8**:

$$OR = \exp(\beta_i) \quad [8]$$

The model results, including estimated coefficients, odds ratios, and 95% confidence intervals, provide a transparent and interpretable framework for understanding how the selected variables influence crash severity outcomes in tree-involved crashes.

## 4. DATA

In this study, we used the Crash Report Sampling System (CRSS) database, spanning from 2020 to 2023. The CRSS is a US nationally representative probability sample of police-reported motor vehicle crashes across 60 specific geographic sites in the US, managed by the National Highway Traffic Safety Administration (NHTSA). The database contains 28 separate data files, covering various features of each crash, such as roadway geometry, driver behavior, vehicle characteristics, and environmental conditions, with approximately 120 data elements extracted from the original police reports.

To analyze tree-involved crashes, a specific dataset is constructed by filtering the Accident file for cases where the First Harmful Event (FHE) was recorded as a tree. Note that, FHE is the initial action or contact in a crash sequence that produces the first injury or property damage. Using the unique identifier 'CASENUM', these crash records were then merged with the corresponding 'Vehicle', 'Person', 'Drimpair', and 'Maneuver' data files to create a comprehensive final dataset, resulting in a sample of 3,778 tree-involved crashes in 4 years period (2020 = 1,061, 2021 = 932, 2022 = 974, 2023 = 811). The annual distribution of crash severity for all crashes and tree-involved crashes in the CRSS database is presented in Figure 2 and Figure 3.

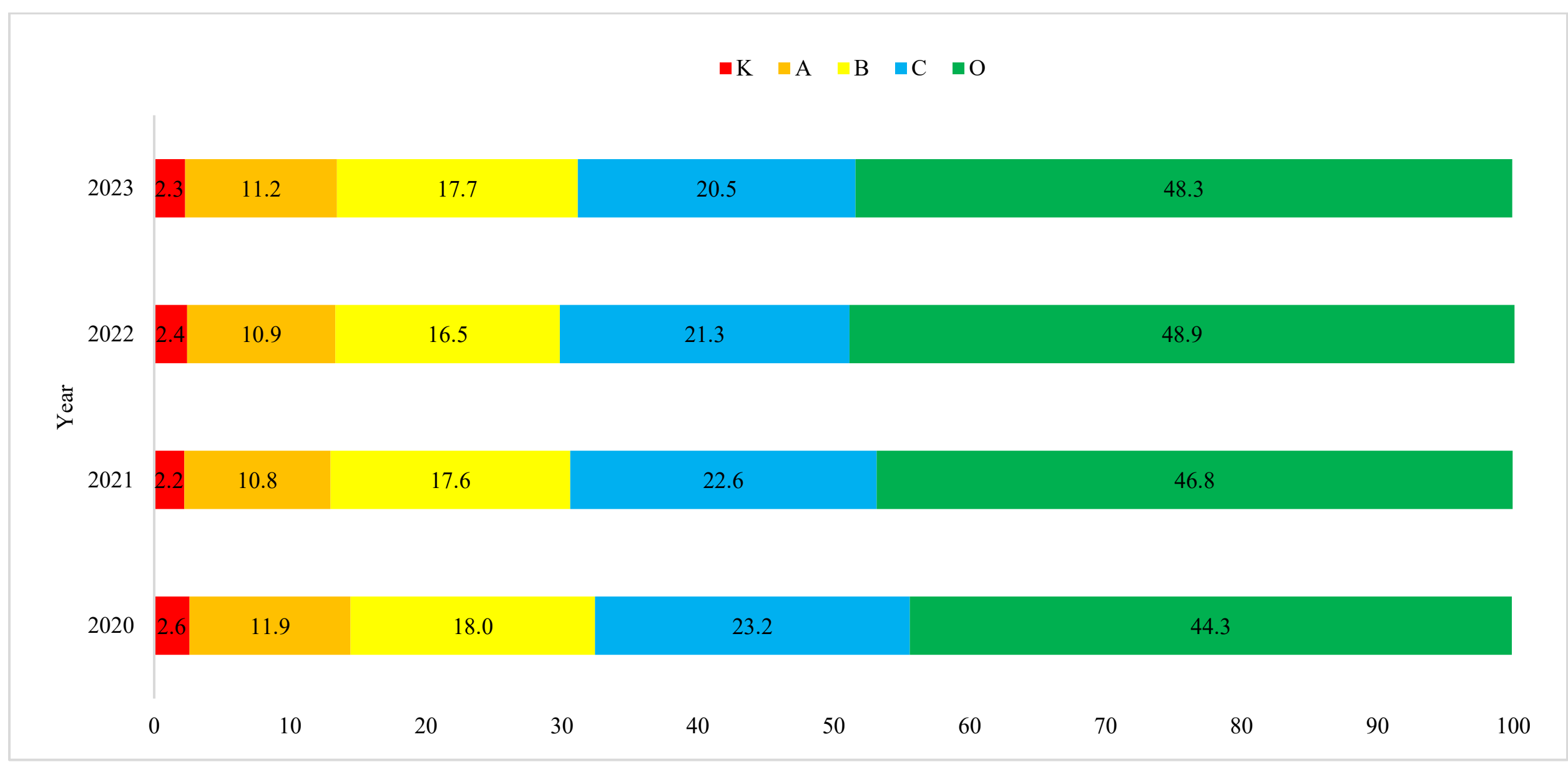

**Figure 2. Annual Percentage Distribution of Crash Severity for All Crashes (2,12,396)**

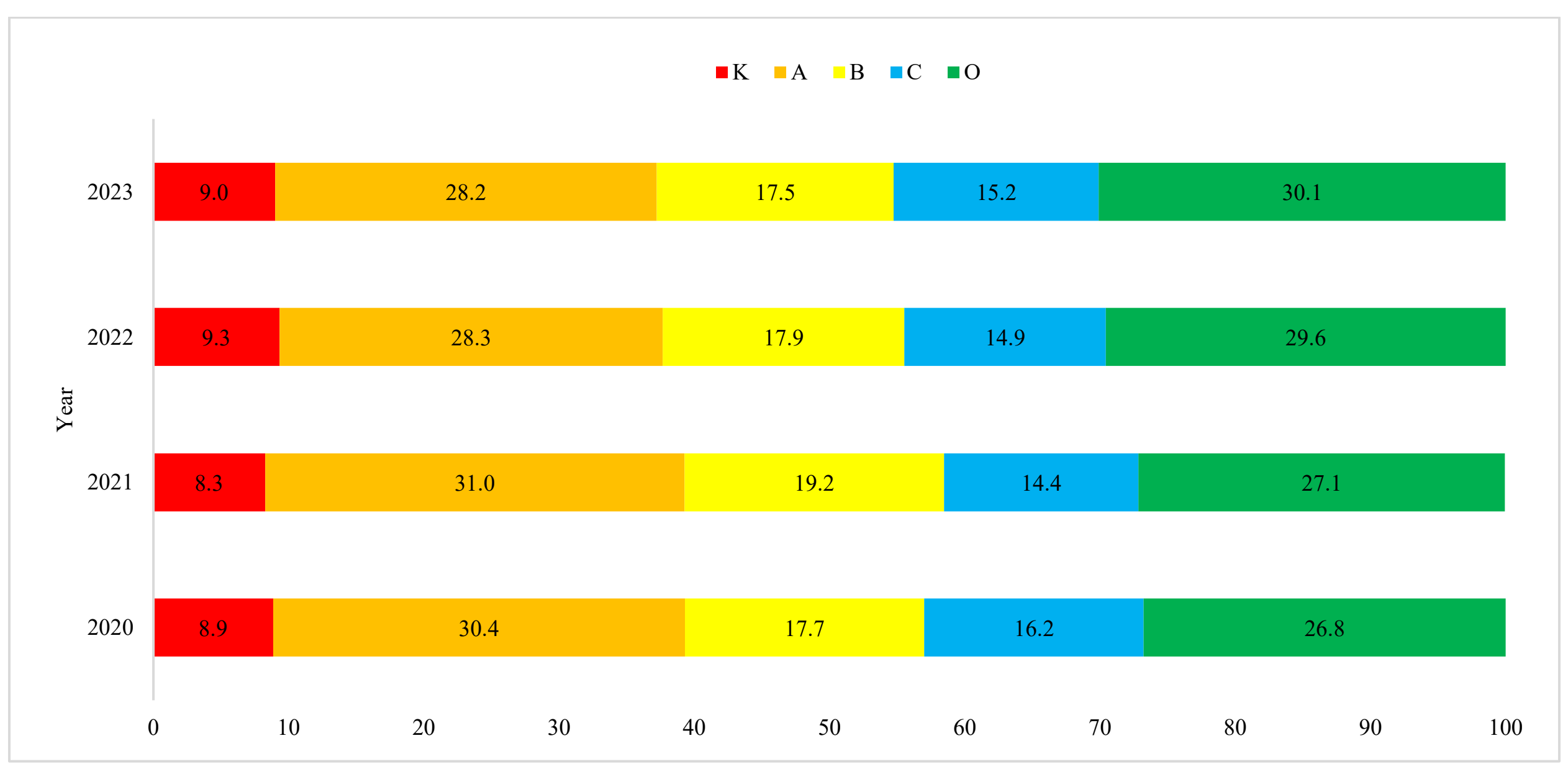

**Figure 3. Annual Percentage Distribution of Crash Severity for Tree-involved Crashes (3,778)**

Overall, the data integration, and analysis flowchart is provided in the following Figure 4.

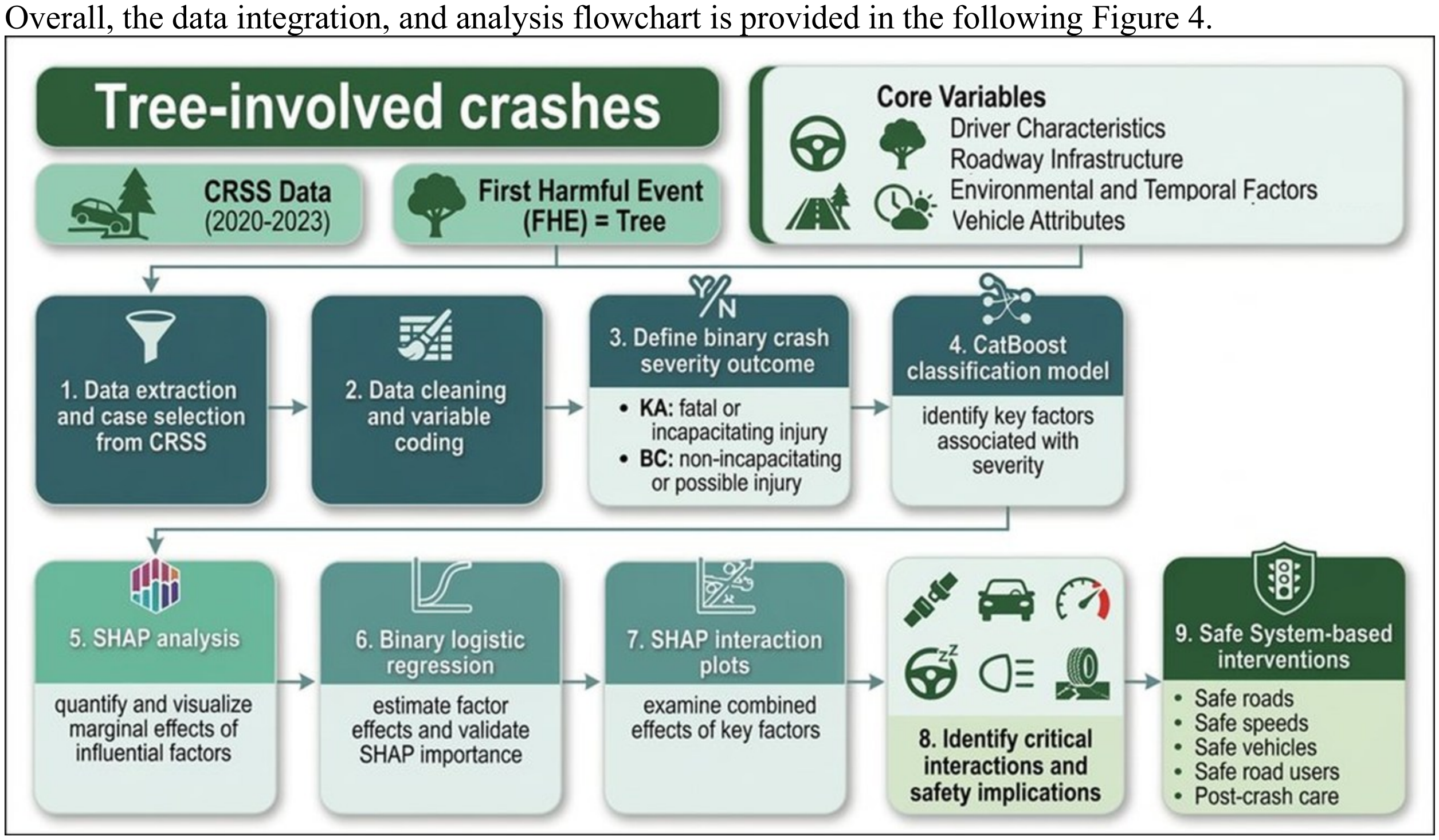

**Figure 4. Data Integration and Analysis Flowchart**

**4.1 Exploratory Data Analysis**

A total of 26 independent variables were primarily selected for this research based on domain knowledge and established findings in safety literature focused on tree-involved crash investigations. To reduce data sparsity and enhance model interpretability, the categories within each variable were recoded into more balanced categories while still maintaining meaningful variable categorization. Crash injury severity is the dependent variable in this study. Originally coded according to the KABCO scale (K: Fatal, A: Suspected Serious Injury, B: Suspected Minor Injury, C: Possible Injury, O: No Apparent Injury), the severity levels were aggregated into a binary classification for analytical purposes: (1) fatal-severe injury (combining K and A) and (2) non-severe injury (combining B and C). Tree-involved crashes resulting in Property Damage Only (PDO) were excluded from the final analysis to focus specifically on injury outcomes. The descriptive statistics and distribution of the independent variables across these two injury severity levels (KA, BC) are also summarized in Table 2.

**Table 2. Descriptive Statistics of Variables in Tree-Involved Crashes**

| Variable | Description | BC (1,257) | KA (1,452) | Total (2,709) |
|---|---|---|---|---|
| **Driver Characteristics** | | | | |
| Age | 0 = 15-24 | 397 (31.58%) | 404 (27.82%) | 801 (29.57%) |
| | 1 = 25-40 | 416 (33.09%) | 544 (37.47%) | 960 (35.44%) |
| | 2 = 41-65 | 292 (23.23%) | 396 (27.27%) | 688 (25.40%) |
| | 3 = 65+ | 152 (12.09%) | 108 (7.44%) | 260 (9.60%) |
| Gender | 0 = Female | 454 (36.12%) | 399 (27.48%) | 853 (31.49%) |
| | 1 = Male | 803 (63.88%) | 1053 (72.52%) | 1856 (68.51%) |
| Alcohol Involved | 0 = Not Involved | 1083 (86.16%) | 1085 (74.72%) | 2168 (80.03%) |
| | 1 = Involved | 174 (13.84%) | 367 (25.28%) | 541 (19.97%) |
| Impairment | 0 = No Impairment | 477 (37.95%) | 369 (25.41%) | 846 (31.23%) |
| | 1 = Unknown | 433 (34.45%) | 564 (38.84%) | 997 (36.80%) |
| | 2 = Physical Impairment | 194 (15.43%) | 214 (14.74%) | 408 (15.06%) |
| | 3 = Alcohol/Drugs | 153 (12.17%) | 305 (21.01%) | 458 (16.91%) |

| Variable | Description | BC (1,257) | KA (1,452) | Total (2,709) |
|---|---|---|---|---|
| Speeding | 0 = No Violation | 909 (72.32%) | 927 (63.84%) | 1836 (67.77%) |
| | 1 = Unknown | 35 (2.78%) | 55 (3.79%) | 90 (3.32%) |
| | 2 = Violated | 313 (24.90%) | 470 (32.37%) | 783 (28.90%) |
| Restraint Use | 0 = Seatbelt Used | 969 (77.09%) | 794 (54.68%) | 1763 (65.08%) |
| | 1 = Unknown | 109 (8.67%) | 149 (10.26%) | 258 (9.52%) |
| | 2 = Not Used | 179 (14.24%) | 509 (35.06%) | 688 (25.40%) |
| Pre-Crash Movement | 0 = Straight/Normal Flow | 704 (56.01%) | 809 (55.72%) | 1513 (55.85%) |
| | 1 = Unknown | 119 (9.47%) | 92 (6.34%) | 211 (7.79%) |
| | 2 = Negotiating Curve | 434 (34.53%) | 551 (37.95%) | 985 (36.36%) |
| Maneuver | 0 = No Evasive Maneuver | 142 (11.30%) | 148 (10.19%) | 290 (10.71%) |
| | 1 = Unknown | 948 (75.42%) | 1212 (83.47%) | 2160 (79.73%) |
| | 2 = Evasive Maneuver Taken | 167 (13.29%) | 92 (6.34%) | 259 (9.56%) |
| **Roadway Infrastructure** | | | | |
| Road Alignment | 0 = Straight | 782 (62.21%) | 856 (58.95%) | 1638 (60.47%) |
| | 1 = Unknown | 14 (1.11%) | 12 (0.83%) | 26 (0.96%) |
| | 2 = Curve | 461 (36.67%) | 584 (40.22%) | 1045 (38.58%) |
| Road Profile | 0 = Level | 830 (66.03%) | 988 (68.04%) | 1818 (67.11%) |
| | 1 = Unknown | 100 (7.96%) | 89 (6.13%) | 189 (6.98%) |
| | 2 = Unlevel | 327 (26.01%) | 375 (25.83%) | 702 (25.91%) |
| Location Type | 0 = Rural | 454 (36.12%) | 551 (37.95%) | 1005 (37.10%) |
| | 1 = Urban | 803 (63.88%) | 901 (62.05%) | 1704 (62.90%) |
| Trafficway | 0 = One-way | 55 (4.38%) | 60 (4.13%) | 115 (4.25%) |
| | 1 = Unknown | 167 (13.29%) | 123 (8.47%) | 290 (10.71%) |
| | 2 = Two-way undivided | 812 (64.60%) | 987 (67.98%) | 1799 (66.41%) |
| | 3 = Two-way divided | 223 (17.74%) | 282 (19.42%) | 505 (18.64%) |
| Speed Limit | 0 = Low Speed | 152 (12.09%) | 125 (8.61%) | 277 (10.23%) |
| | 1 = Unknown | 128 (10.18%) | 150 (10.33%) | 278 (10.26%) |
| | 2 = Medium Speed | 656 (52.19%) | 702 (48.35%) | 1358 (50.13%) |
| | 3 = High Speed | 321 (25.54%) | 475 (32.71%) | 796 (29.38%) |
| Intersection | 0 = Intersection | 120 (9.55%) | 80 (5.51%) | 200 (7.38%) |
| | 1 = Unknown | 13 (1.03%) | 7 (0.48%) | 20 (0.74%) |
| | 2 = Not an intersection | 1124 (89.42%) | 1365 (94.01%) | 2489 (91.88%) |
| Traffic Control | 0 = No Controls | 887 (70.56%) | 1086 (74.79%) | 1973 (72.83%) |
| | 1 = Unknown | 290 (23.07%) | 285 (19.63%) | 575 (21.23%) |
| | 2 = Traffic Controls | 80 (6.36%) | 81 (5.58%) | 161 (5.94%) |
| **Environmental and Temporal Factors** | | | | |
| Month | 0 = Spring | 270 (21.48%) | 327 (22.52%) | 597 (22.04%) |
| | 1 = Summer | 318 (25.30%) | 401 (27.62%) | 719 (26.54%) |
| | 2 = Autumn | 333 (26.49%) | 378 (26.03%) | 711 (26.25%) |
| | 3 = Winter | 336 (26.73%) | 346 (23.83%) | 682 (25.18%) |
| Hour | 0 = 06:00-12:00 | 295 (23.47%) | 259 (17.84%) | 554 (20.45%) |
| | 1 = 12:00-18:00 | 373 (29.67%) | 361 (24.86%) | 734 (27.09%) |
| | 2 = 18:00-00:00 | 327 (26.01%) | 447 (30.79%) | 774 (28.57%) |
| | 3 = 00:00-06:00 | 262 (20.84%) | 385 (26.52%) | 647 (23.88%) |
| Weather | 0 = Clear | 1001 (79.63%) | 1240 (85.40%) | 2241 (82.72%) |
| | 1 = Adverse | 256 (20.37%) | 212 (14.60%) | 468 (17.28%) |
| Lighting | 0 = DayLight | 660 (52.51%) | 645 (44.42%) | 1305 (48.17%) |
| | 1 = Dark Lighted | 275 (21.88%) | 326 (22.45%) | 601 (22.19%) |
| | 2 = Dark Not Lighted | 322 (25.62%) | 481 (33.13%) | 803 (29.64%) |
| Rush Hour | 0 = No | 916 (72.87%) | 1117 (76.93%) | 2033 (75.05%) |
| | 1 = Yes (06:00 – 09:00 and 16:00 – 19:00) | 341 (27.13%) | 335 (23.07%) | 676 (24.95%) |

| Variable | Description | BC (1,257) | KA (1,452) | Total (2,709) |
| --- | --- | --- | --- | --- |
| Weekend | 0 = Weekdays | 871 (69.29%) | 908 (62.53%) | 1779 (65.67%) |
| | 1 = Weekends | 386 (30.71%) | 544 (37.47%) | 930 (34.33%) |
| Road Surface | 0 = Dry | 864 (68.74%) | 1114 (76.72%) | 1978 (73.02%) |
| | 1 = Unknown | 35 (2.78%) | 26 (1.79%) | 61 (2.25%) |
| | 2 = Wet/Adverse | 358 (28.48%) | 312 (21.49%) | 670 (24.73%) |
| **Vehicle Attributes** | | | | |
| Impact Point | 0 = Front | 1059 (84.25%) | 1192 (82.09%) | 2251 (83.09%) |
| | 1 = Unknown | 39 (3.10%) | 50 (3.44%) | 89 (3.29%) |
| | 2 = Back | 115 (9.15%) | 184 (12.67%) | 299 (11.04%) |
| | 3 = Side | 44 (3.50%) | 26 (1.79%) | 70 (2.58%) |
| Vehicle Type | 0 = Passenger Car | 605 (48.13%) | 693 (47.73%) | 1298 (47.91%) |
| | 1 = Other/Unknown | 75 (5.97%) | 58 (3.99%) | 133 (4.91%) |
| | 2 = Motorcycle | 32 (2.55%) | 48 (3.31%) | 80 (2.95%) |
| | 3 = SUV | 356 (28.32%) | 378 (26.03%) | 734 (27.09%) |
| | 4 = Truck/Tractor | 189 (15.04%) | 275 (18.94%) | 464 (17.13%) |
| Vehicle Age | 0 = Newer | 466 (37.07%) | 353 (24.31%) | 819 (30.23%) |
| | 1 = Middle | 248 (19.73%) | 292 (20.11%) | 540 (19.93%) |
| | 2 = Older | 543 (43.20%) | 807 (55.58%) | 1350 (49.83%) |
| Airbag Deployed | 0 = Not Deployed | 438 (34.84%) | 415 (28.58%) | 853 (31.49%) |
| | 1 = Deployed Front | 226 (17.98%) | 273 (18.80%) | 499 (18.42%) |
| | 2 = Deployed Combination | 308 (24.50%) | 431 (29.68%) | 739 (27.28%) |
| | 3 = Deployed Unknown Location | 285 (22.67%) | 333 (22.93%) | 618 (22.81%) |

The descriptive statistics reveal distinct patterns between KA and BC injury in tree-involved crashes across multiple dimensions. Male drivers accounted for a larger share of KA crashes (72.52%) than of BC crashes (63.88%). Drivers aged 25-40 years were overrepresented in KA outcomes, while those aged 65 years and older were underrepresented. Unsafe driving behaviors were strongly associated with severity. For example, alcohol involvement was approximately 1.8 times higher in KA crashes, and alcohol or drug impairment demonstrated a similar elevation. Speeding violations occurred more frequently in KA crashes (32.37% vs. 24.90%). Seatbelt usage showed the most striking difference: around 35% of unrestrained drivers were involved in KA injuries (compared to 14.24% in BC injuries), highlighting the critical protective value of restraints in tree-involved collisions. Environmental and temporal factors exhibited consistent patterns with crash severity. Dark conditions without street lighting are associated with 33.13% of KA crashes versus 25.62% of BC crashes, suggesting the critical role of lighting conditions in crash severity in tree-involved collisions. Weekend occurrences and late-night hours (00:00–06:00) were also more common in KA crashes.

## 5. RESULTS AND DISCUSSION

### 5.1 Cat Boost Model

The CatBoost classification model was developed using an 80:20 train-test split with stratified sampling to maintain class distribution. The model was configured with a depth of 6, 1,000 iterations, a learning rate of 0.1, and a Log loss function. Early stopping was implemented with a patience of 50 iterations to prevent overfitting. The CatBoost model achieved a test set accuracy of 0.68 and demonstrated balanced performance across both injury severity classes. For the KA injury class, the model achieved precision of 0.71, recall of 0.70, and F1-score of 0.71, while the BC severe injury class yielded precision of 0.66, recall of 0.66, and F1-score of 0.66. The model converged at iteration 64, with early stopping preventing performance degradation. Rather than optimizing predictive performance through data balancing techniques such as oversampling or SMOTE, this study intentionally preserved the empirical structure of real-world crash data. The CatBoost model served as a feature discovery and pattern recognition tool, with the primary objective of identifying influential factors and their interactions. This machine learning component complements the statistical modeling framework, enabling robust interpretation through SHAP analysis while maintaining data integrity for rigorous causal inference.

**5.2 SHAP-Based Feature Importance Analysis**

Based on the binary classifier (KA vs. BC), the top factors were identified based on the CatBoost model. Next, a SHAP plot (Figure 5) is developed to identify how each factor contributes to tree-involved crash severity.

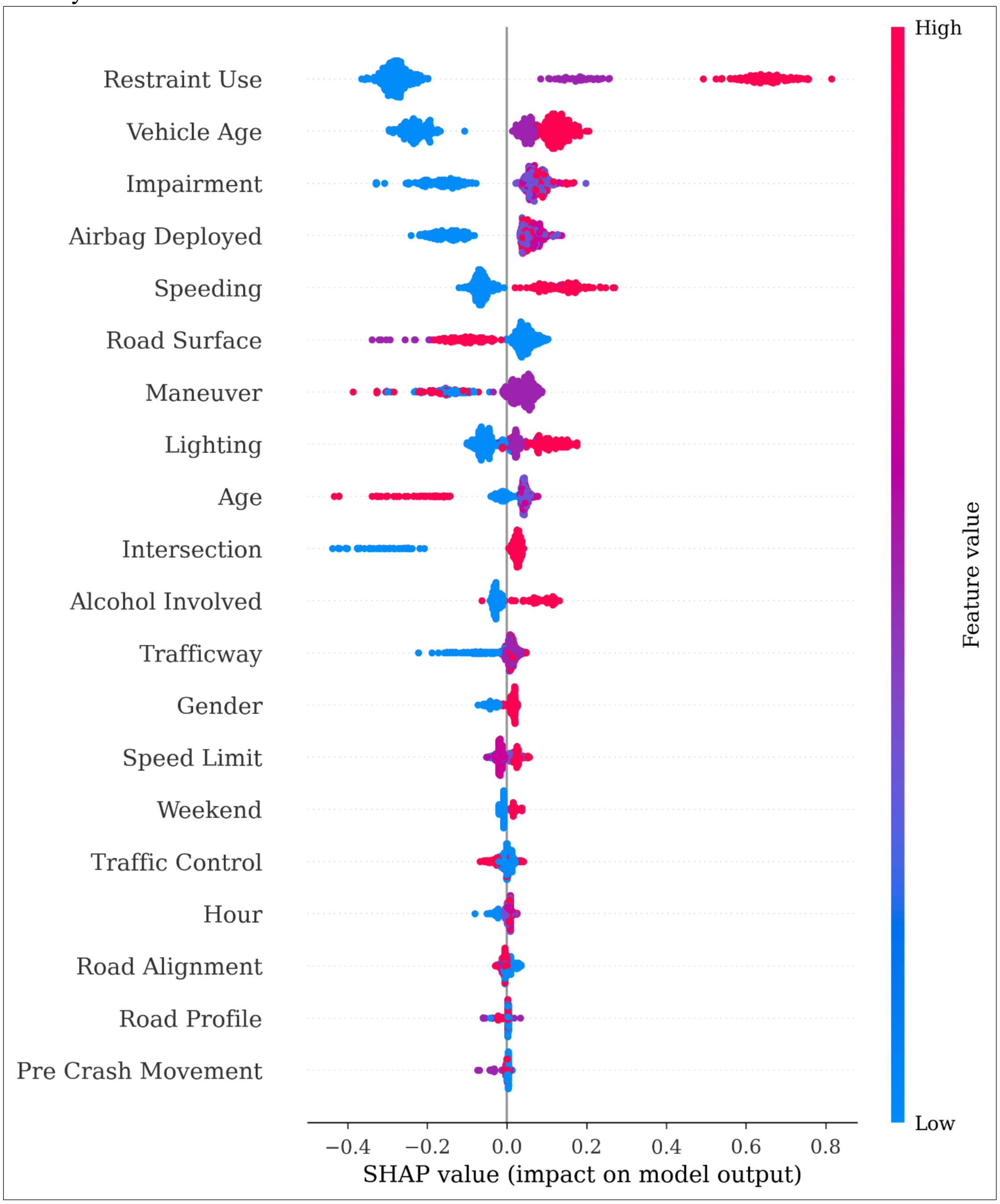

**Figure 5. SHAP Beeswarm Plot for Top Factors**

#### 5.2.1 *Restraint Use*

Restraint use (topmost feature) emerges as one of the most influential predictors of crash severity in tree-involved collisions, exhibiting SHAP values ranging from approximately -0.4 to +0.8. When seatbelts are not used (feature value = 2), the model generates strongly positive SHAP values (approximately +0.6 to +0.8), indicating a substantial increase in the probability of KA outcomes, while seatbelt use (feature value = 0) produces negative SHAP values (approximately -0.3 to -0.4), suggesting a protective effect that shifts predictions toward BC outcomes. This strong effect reflects the unique hazards of ROR tree crashes, in which vehicles experience rapid deceleration and multiple impact events, including loss of control, potential rollover, and high-energy collisions with a tree. Unrestrained occupants are at significantly higher risk of ejection from the vehicle or severe contact with the interior and exterior environment during these sequential impacts, dramatically increasing the likelihood of KA injuries. This finding is consistent with previous research emphasizing the life-saving impact of seatbelt usage [53–55]. As such, properly worn seatbelts may mitigate these risks by maintaining occupant position, distributing crash forces across stronger body structures, preventing ejection, and allowing supplemental restraint systems to function as designed.

#### 5.2.2*Vehicle Age*

Vehicle age demonstrates a moderate but notable influence on crash severity in tree-involved collisions, with SHAP values ranging from approximately -0.3 to +0.3. Older vehicles (feature value = 2, shown in red) cluster toward positive SHAP values, indicating increased likelihood of KA crashes, while newer vehicles (feature value = 0, shown in blue) tend toward negative SHAP values, suggesting BC injury outcomes. This pattern likely reflects the substantial safety advancements in modern vehicle design, including improved structural crashworthiness, enhanced crumple zones that absorb impact energy, sophisticated restraint systems, and additional safety features such as electronic stability control that can prevent ROR departures altogether. Older vehicles lack these protective technologies and may also exhibit degraded structural integrity and worn safety equipment, making occupants more vulnerable during high-energy tree-involved collisions. This result corroborates previous studies highlighting the safety advantages of newer vehicle models over older ones [56–58].

#### 5.2.3 *Airbag Deployment*

Airbag deployment shows a complex relationship with crash severity outcome in tree-involved collisions, with SHAP values ranging from approximately -0.2 to +0.3. Counterintuitively, airbag deployment (feature values 1, 2, and 3, shown in red) is associated with positive SHAP values, indicating an increased likelihood of KA outcomes, whereas non-deployment (feature value = 0, shown in blue) is associated with negative SHAP values, suggesting BC outcomes. This pattern does not imply that airbags cause more severe injuries; rather, airbag deployment serves as a marker of crash severity itself - airbags deploy only when crash sensors detect sufficient deceleration force, meaning their deployment indicates that the vehicle experienced a high-energy impact with the tree, which inherently carries greater risk of severe injury regardless of the airbag's protective effect. This aligns with previous research, which treats airbag deployment as a proxy for high-severity crashes [59,60].

#### 5.2.4 *Driver Impairment and Involvement*

Driver impairment has a strong influence on crash severity in tree-involved collisions, with SHAP values spanning approximately -0.3 to +0.3. Alcohol/drug impairment (feature value = 3, shown in red) and physical impairment (feature value = 2) cluster toward positive SHAP values, indicating substantially increased likelihood of KA outcomes, while no impairment (feature value = 0, shown in blue) is associated with negative SHAP values, suggesting BC crashes. Impaired drivers are more likely to experience severe crashes because impairment affects their ability to perceive roadway hazards, maintain vehicle control, take corrective action before leaving the roadway, and reduce speed before impact, resulting in higher-energy collisions with trees and reduced capacity to protect themselves during the crash sequence.

Alcohol involvement demonstrates a notable influence on crash severity in tree-involved collisions, with SHAP values ranging from approximately -0.2 to +0.3. Crashes involving alcohol (feature value = 1, shown in red) are associated with positive SHAP values, suggesting a higher probability of KA crashes, whereas crashes without alcohol involvement (feature value = 0, shown in blue) are associated with negative SHAP values, suggesting BC outcomes. This pattern reflects the impairing effects of alcohol on critical driving functions, including judgment, reaction time, visual processing, and motor coordination, which increase the likelihood that impaired drivers fail to recognize hazardous conditions, maintain proper lane position, or take effective corrective action when the vehicle begins to leave the travel lanes. Additionally, alcohol-involved crashes often occur at higher speeds due to reduced risk perception and impaired speed regulation, resulting in more violent impacts with trees and greater occupant injury severity during the rapid deceleration of fixed-object collisions. The elevated crash severity observed for impaired drivers in this study is consistent with prior research showing that alcohol and impairment substantially increase the likelihood of serious and fatal injuries in motor vehicle crashes from a general perspective [38,61,62].

#### 5.2.5 *Speeding*

Speeding violations demonstrate a substantial impact on crash severity in tree-involved collisions, with SHAP values ranging from approximately -0.3 to +0.5. Speed violations (feature value = 2, shown in red) are strongly associated with positive SHAP values, indicating a significantly increased probability of KA outcomes, whereas no speed violations (feature value = 0, shown in blue) are associated with negative SHAP values, suggesting BC outcomes. This relationship reflects fundamental crash physics - higher speeds exponentially increase kinetic energy, resulting in greater impact forces during tree collisions, longer braking distances that prevent speed reduction before impact, reduced driver reaction time to navigate back onto the roadway, and more severe occupant injuries due to the violent deceleration forces experienced during high-speed fixed-object impacts. The adverse effects of speeding have been widely reported in earlier studies of crash severity [63–65].

#### 5.2.6 *Lighting Condition*

Lighting conditions show a moderate influence on crash severity in tree-involved collisions, with SHAP values ranging from approximately -0.3 to +0.3. Dark conditions, whether lighted (feature value = 1) or not lighted (feature value = 2), shown in red, are associated with positive SHAP values indicating increased likelihood of KA outcomes, while daylight conditions (feature value = 0, shown in blue) correspond to negative SHAP values suggesting BC outcomes. Reduced visibility in dark conditions impairs drivers' ability to perceive roadway hazards, roadway edges, and obstacles early enough to take corrective action, increasing the likelihood that vehicles depart the roadway at higher speeds. As such, drivers are unable to brake or steer effectively before striking trees, resulting in higher-energy impacts leading to severe injuries. Previous research has similarly emphasized the adverse safety impacts of reduced visibility under dark conditions [66–68]. In contrast, daylight conditions enhance visual perception and situational awareness, allowing drivers to better recognize hazards, respond earlier, and reduce impact severity, which contributes to lower injury severity in tree-involved crashes.

#### 5.2.7 *Intersection*

Intersection location shows a modest influence on crash severity in tree-involved collisions, with SHAP values ranging from approximately -0.3 to +0.2. Non-intersection locations (feature value = 2, shown in red) are associated with positive SHAP values, indicating an increased likelihood of KA crashes, while intersection locations (feature value = 0, shown in blue) are associated with negative SHAP values, suggesting BC outcomes. This pattern largely reflects where trees are physically present in the roadway environment. Trees are rarely planted at or near intersections and are far more commonly found along roadside segments between intersections. Consequently, most tree-involved crashes occur at non-intersection locations, where higher travel speeds result in higher-speed departures from the roadway and more severe tree impacts. The apparent protective association of intersection locations should therefore be interpreted cautiously, as it may reflect the near absence of roadside trees at those locations rather than any inherent safety benefit of intersections themselves. Similar findings have been documented in previous research [40,69,70].

**5.2.8** ***Road Surface Condition***

Road surface conditions demonstrate a moderate influence on crash severity in tree-involved collisions, with SHAP values ranging from approximately -0.4 to +0.2. Wet or adverse road surfaces (feature value = 2, shown in red) are associated with positive SHAP values, indicating an increased likelihood of KA injuries, while dry surfaces (feature value = 0, shown in blue) are associated with negative SHAP values, suggesting BC outcomes. This relationship reflects how wet road surfaces reduce tire-pavement friction and vehicle stability, impairing drivers' ability to maintain control, execute emergency maneuvers, and brake effectively when attempting to avoid obstacles. Reduced traction on wet or adverse surfaces increases the likelihood of vehicles departing the roadway and makes it more difficult for drivers to slow before impact, resulting in higher-energy collisions with trees. This finding aligns with prior injury severity studies that identify wet or adverse road surface conditions as being associated with higher crash injury severity [71–73].

**5.3 Binary Logistic Regression Model Results**

In this step, the first set of predictor variables is checked for multicollinearity. After this step, the binary logistic regression model is applied to predict crash severity (KA vs. BC).

**5.3.1** ***Multicollinearity Check***

The results obtained from Cramer's V are provided in Figure 6**.** Three variables showing high correlation (Cramér's V > 0.50) were removed. 'Impairment' variable was retained over 'alcohol involved' (V = 0.77), 'lighting condition' variable was retained over 'hour' (V = 0.57), and 'road alignment' variable was retained over 'pre-crash movement' (V = 0.68).

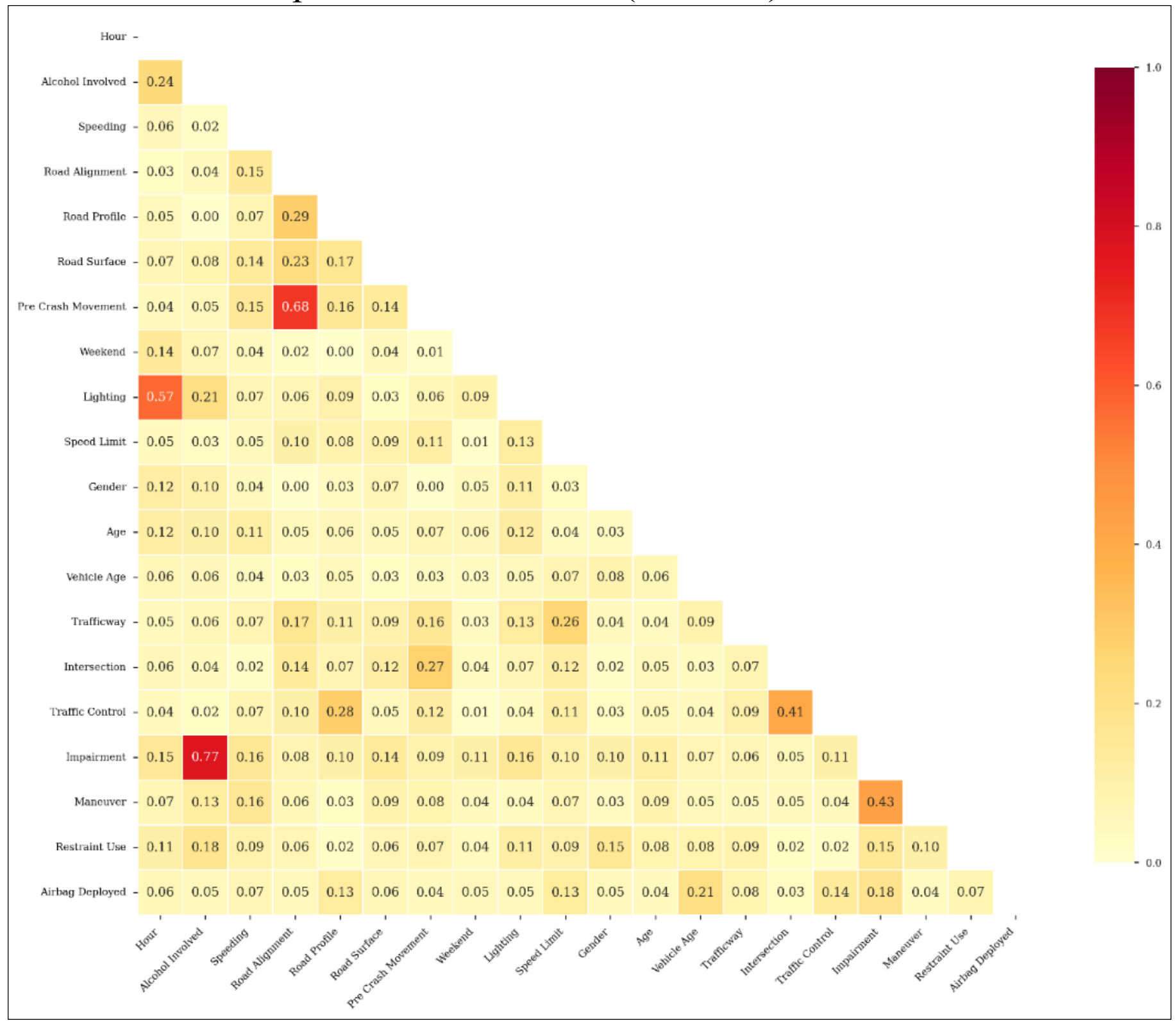

**Figure 6. Cramer's V Statistics**

Based on the final set of variables (total = 23), a binary logistic regression model is used to predict the likelihood of KA injury (BC injury as the reference) in tree-involved collisions. Note that the reference level for each variable category is selected based on domain knowledge and the intended exploration of crash factors. For example, 'seatbelt used' is used as a reference to quantify the effect of 'restraint not used'. Overall, the binary logistic regression model results are shown in Table 3.

**Table 3. Binary Logistic Regression Model Results**

| Variable | Coeff. | S.E. | z value | p-value | OR |
|---|---|---|---|---|---|
| Constant | -3.76 | 0.47 | -7.99 | <0.0001 | - |
| Restraint Use = Not Used *(ref: Seatbelt Used)* | 1.06 | 0.12 | 8.86 | <0.0001 | 2.90 |
| Vehicle Age = Older *(ref: Newer)* | 0.82 | 0.12 | 6.98 | <0.0001 | 2.27 |
| Airbag Deployed = Deployed Combination *(ref: Not Deployed)* | 0.76 | 0.13 | 5.65 | <0.0001 | 2.14 |
| Vehicle Age = Middle *(ref: Newer)* | 0.52 | 0.14 | 3.78 | 0.0002 | 1.68 |
| Airbag Deployed = Deployed Front *(ref: Not Deployed)* | 0.54 | 0.15 | 3.68 | 0.0002 | 1.72 |
| Speed Limit = Unknown *(ref: Low Speed)* | 0.82 | 0.23 | 3.61 | 0.0003 | 2.26 |
| Speeding = Violated *(ref: No Violation)* | 0.40 | 0.12 | 3.51 | 0.0004 | 1.50 |
| Impairment = Alcohol/Drugs *(ref: No Impairment)* | 0.54 | 0.15 | 3.50 | 0.0005 | 1.71 |
| Impairment = Unknown *(ref: No Impairment)* | 0.41 | 0.12 | 3.33 | 0.0009 | 1.51 |
| Intersection = Not an intersection *(ref: Intersection)* | 0.81 | 0.24 | 3.32 | 0.0009 | 2.24 |
| Speed Limit = High Speed *(ref: Low Speed)* | 0.58 | 0.18 | 3.24 | 0.0012 | 1.79 |
| Age_41-65 *(ref: 65+)* | 0.58 | 0.18 | 3.17 | 0.0015 | 1.78 |
| Airbag Deployed = Deployed Unknown Location *(ref: Not Deployed)* | 0.40 | 0.14 | 2.95 | 0.0032 | 1.49 |
| Restraint Use = Unknown *(ref: Seatbelt Used)* | 0.44 | 0.17 | 2.64 | 0.0082 | 1.55 |
| Lighting = Dark Not Lighted *(ref: DayLight)* | 0.30 | 0.12 | 2.62 | 0.0087 | 1.35 |
| Impairment = Physical Impairment *(ref: No Impairment)* | 0.47 | 0.18 | 2.58 | 0.0098 | 1.61 |
| Road Surface = Dry *(ref: Wet/Adverse)* | 0.29 | 0.11 | 2.58 | 0.0100 | 1.34 |
| Driver Age = 25-40 *(ref: 65+)* | 0.45 | 0.18 | 2.52 | 0.0116 | 1.56 |
| Weekend = Weekends *(ref: Weekdays)* | 0.24 | 0.10 | 2.35 | 0.0187 | 1.27 |

**Note: only significant variables are reported in the table*

The binary logistic regression model identifies restraint use as the most influential predictor of KA crashes in tree-involved collisions. Non-use of seatbelts yields an OR of 2.9 (p<0.0001), indicating that unrestrained occupants are nearly three times more likely to sustain fatal or incapacitating injuries in tree-involved collisions than those wearing seatbelts. This finding aligns closely with the SHAP analysis, which showed that restraint non-use had the strongest positive impact on severity predictions, with SHAP values reaching +0.8. Vehicle age also emerges as significant in both models, with older vehicles showing an OR of 2.27 (p<0.0001) in the logistic regression model and corresponding positive SHAP values in the SHAP plot, reflecting the reduced crashworthiness and safety features in aging vehicles. Similarly, speeding violations (OR = 1.5, p = 0.0004) and alcohol/drug impairment (OR = 1.71, p = 0.0005) are significant predictors in the logistic regression model and demonstrate substantial positive SHAP values, confirming that higher speeds and impaired driving consistently increase crash severity through increased impact energy and reduced driver control during ROR events.

Interestingly, airbag deployment presents consistent patterns across both analytical approaches but requires careful interpretation. In the logistic regression model, deployed airbags show elevated OR, combination deployment (OR = 2.14, p<0.0001), front deployment (OR = 1.72, p = 0.0002), and unknown location deployment (OR = 1.49, p = 0.0032), while the SHAP plot similarly shows positive SHAP values for all deployment categories. Both models confirm that airbag deployment serves as a severity marker rather than a causal factor, as airbags deploy only when crash sensors detect high-energy impacts, indicating the crash was already severe enough to trigger deployment thresholds. Lighting conditions show convergence between models, with dark not lighted conditions (OR = 1.35, p = 0.0087) associated with increased severity in the logistic regression model, mirroring the positive SHAP values for dark conditions in the SHAP analysis.

The two modeling approaches reveal few differences in variable importance rankings and effect magnitudes. Intersection location demonstrates statistical significance in the logistic model (non-intersection OR = 2.24, p = 0.0009) and shows moderate SHAP values, with both indicating that tree-involved collisions at midblock locations are more severe due to higher speeds and fewer traffic controls. Driver age categories show interesting patterns, with middle-aged drivers (41-65 years, OR = 1.78, p = 0.0015) and comparatively young drivers (25-40 years, OR=1.56, p=0.0116) showing elevated risk in the logistic model, though the SHAP plot shows more modest effects for age with substantial overlap between categories. Road surface conditions (dry, OR = 1.34, p = 0.01) and weekend (OR = 1.56, p = 0.0116) are statistically significant in the regression model, though their SHAP values suggest more moderate, context-dependent effects. Overall, both methods consistently identify restraint use, vehicle age, speeding, and impairment as the primary drivers of crash severity in tree-involved collisions, with the SHAP analysis providing additional insight into the distribution and interaction effects that the logistic regression coefficients alone cannot capture.

### 5.4 SHAP Interaction Plots

Based on SHAP and a binary logistic regression model, this study also investigated interactions among the most influential factors using the SHAP interaction plot. The goal of this investigation is to quantify the combined effect of multiple factors on crash severity in tree-involved collisions. After several trials of combinations, the following meaningful interactions were identified and selected for further investigations:

a) Lighting condition × Vehicle age
b) Road surface condition × Speeding
c) Speeding × Lighting condition
d) Vehicle age × Restraint use

#### 5.4.1 Lighting Condition and Vehicle Age

The SHAP interaction plot showing the combined effect of lighting condition and vehicle age on tree-involved crash severity is illustrated in Figure 7.

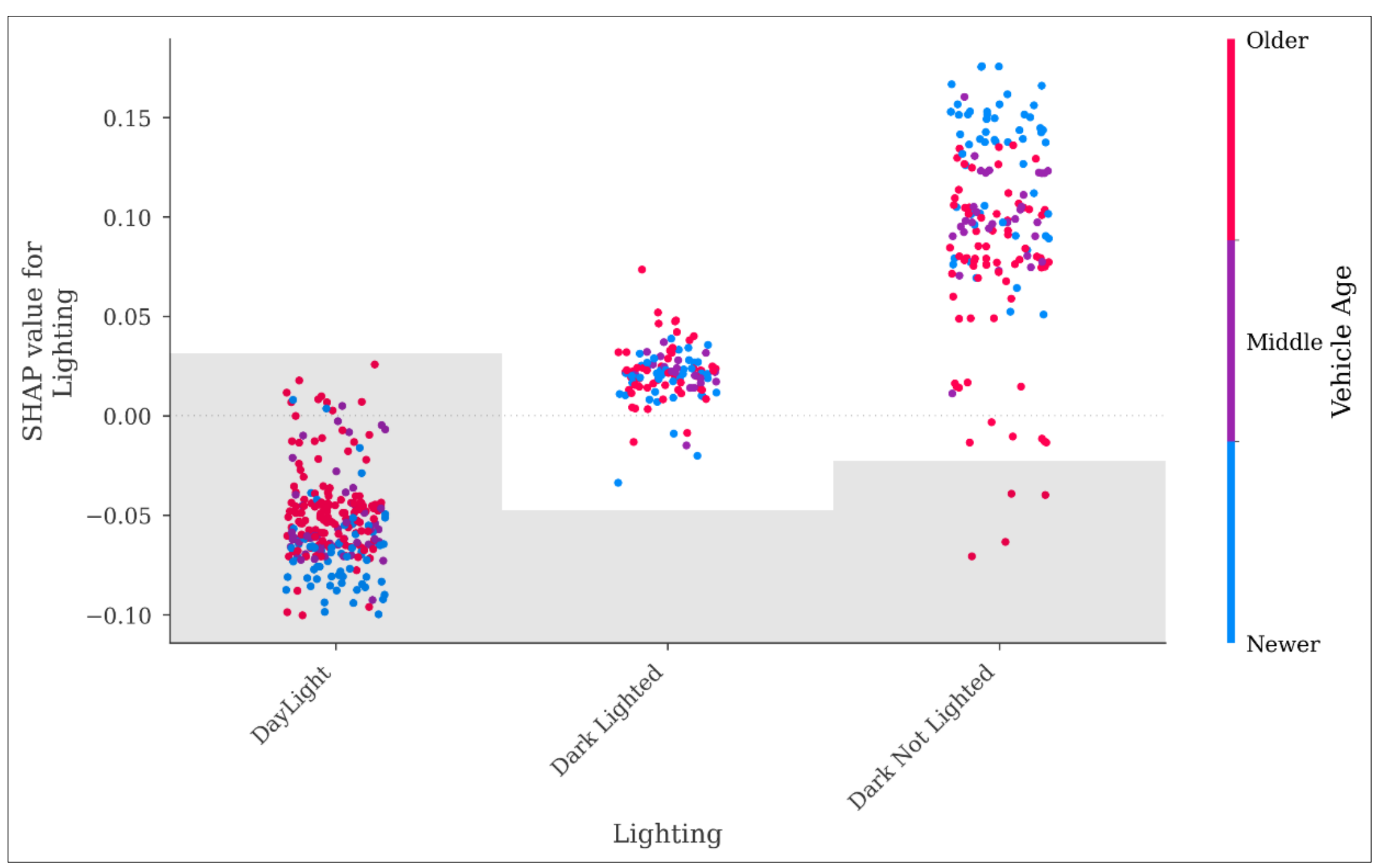

**Figure 7. SHAP Interaction Plot for Lighting Condition and Vehicle Age**

The SHAP interaction plot reveals a strong, nonlinear, and conditional effect of lighting on tree-involved crashes across different vehicle age groups. Under daylight conditions, vehicle age shows minimal differentiation, with newer, middle-aged, and older vehicles clustering together around slightly negative SHAP values (-0.05 to -0.10), indicating that when visibility is optimal, drivers can clearly perceive roadway edges, curves, and potential departure hazards regardless of vehicle age, allowing them to maintain lane position or take corrective action before leaving the roadway. However, the interaction becomes pronounced in dark-not-lighted conditions where newer vehicles (shown in blue) demonstrate the highest positive SHAP values (+0.10 to +0.17), exceeding those of older vehicles (shown in red, ranging from +0.05 to +0.14). This counterintuitive pattern may reflect behavioral compensation effects. Drivers of newer vehicles equipped with superior headlamps, advanced driver assistance systems, and enhanced safety features may exhibit riskier driving behaviors in dark conditions, including higher speeds and reduced caution, under a false sense of security that their vehicle technology will protect them [74]. When ROR departures occur in dark-not-lighted conditions at high speeds, even the advanced safety features of newer vehicles cannot fully mitigate the extreme forces of high-energy tree impacts, while older vehicle occupants may drive more cautiously in poor visibility conditions due to awareness of their vehicles' limitations, potentially reducing pre-crash speeds and resulting in somewhat lower severity despite inferior safety equipment.

### 5.4.2 Road Surface Condition and Speeding

The SHAP interaction plot showing the combined effect of road surface condition and speeding on tree-involved crash severity is depicted in Figure 8.

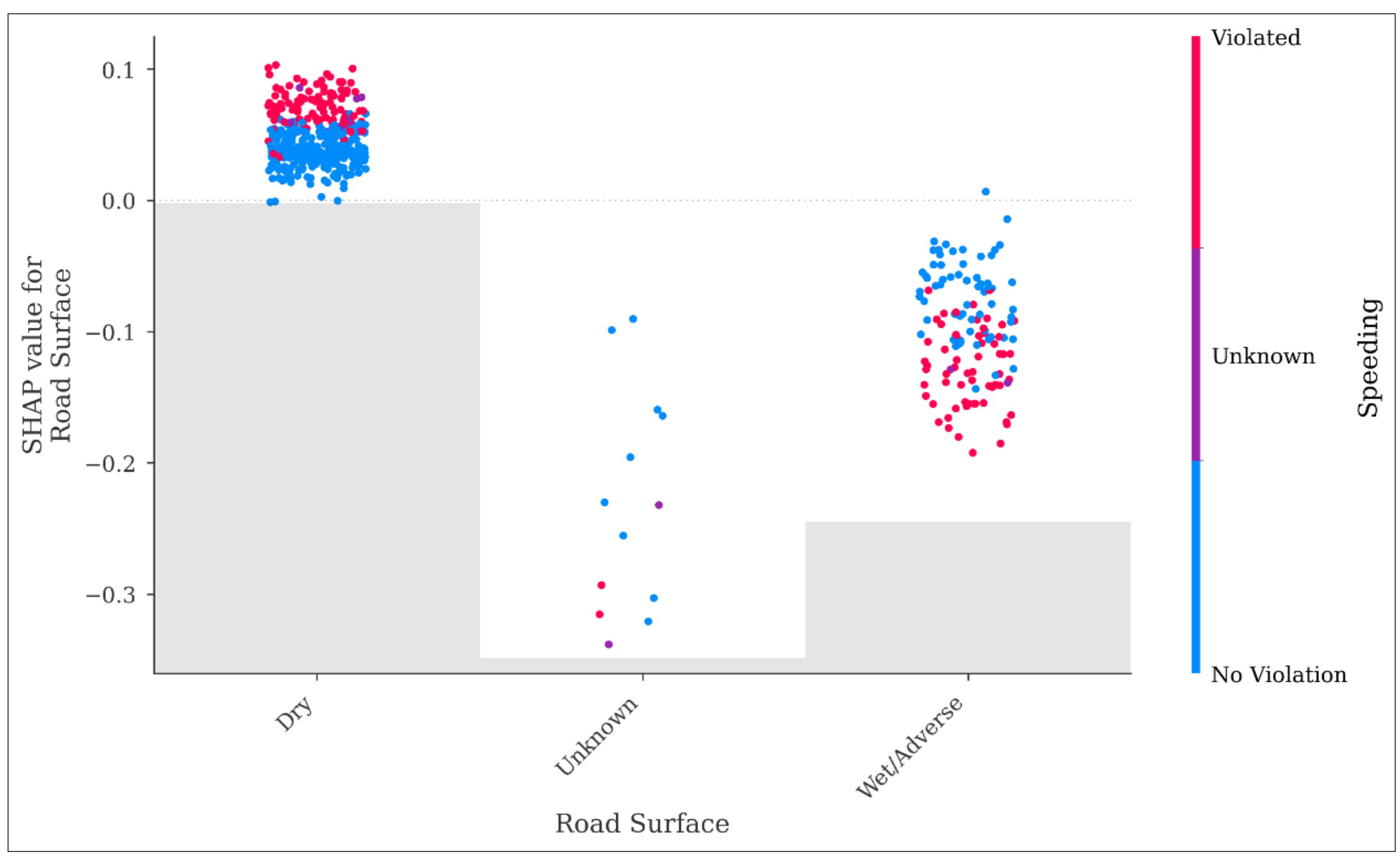

**Figure 8. SHAP Interaction Plot for Road Surface Condition and Speeding**

This SHAP interaction plot illustrates how the effect of road surface conditions on crash severity is moderated by speeding behavior in tree-involved collisions. On dry road surfaces, speeding violations (shown in red) and no violations (shown in blue) cluster tightly together with positive SHAP values ranging from approximately +0.02 to +0.10, indicating that when road conditions are optimal, speeding has relatively modest additional impact on severity beyond the baseline risk. The density of observations is notably highest here, reflecting that the majority of tree-involved collisions in the dataset occur on dry surfaces. Both speeding and non-speeding crashes receive similarly positive SHAP contributions, suggesting that the road surface condition itself is the dominant driver of severity prediction in this group rather than speeding status.

On wet/adverse road surfaces, both speeding violators (red) and non-violators (blue) exhibit negative SHAP values ranging approximately from -0.05 to -0.20. Unlike what might be expected, there is no pronounced divergence between the two speeding groups on wet surfaces - both are distributed across a similar negative range, with moderate intermingling of red and blue dots. This suggests that on wet/adverse surfaces, the road condition again dominates the interaction, with speeding status providing limited additional differentiation in the model's severity predictions.

While counterintuitive, the negative SHAP values observed for wet/adverse surfaces likely reflect a complex interplay of behavioral and definitional factors. A critical starting point is recognizing that the speeding violation category is a broad, heterogeneous construct. A driver traveling 35 mph in a 30-mph zone and a driver traveling 50 mph in a 30-mph zone are both captured under the same binary speeding flag, yet their crash mechanics, injury potential, and likelihood of fatality differ dramatically. This definitional compression means the model is learning from a highly mixed signal. The speeding category likely contains a large proportion of marginal speeders whose excess velocity is modest, which would naturally attenuate the apparent severity effect and could explain why speeding on wet roads does not produce the strongly positive SHAP values one might expect.

Additionally, SHAP values here reflect the model’s learned associations between feature combinations and the severity outcome as coded in the dataset. If severe fatalities on wet roads are underrepresented due to reporting biases, scene clearance practices, or differences in how severity is operationalized across injury thresholds, the model may paradoxically assign lower severity ratings to these conditions. Selection effects may further compound this - crashes on wet surfaces at speed may more frequently involve ROR trajectories, in which the tree becomes a secondary contact after the primary loss of control has already absorbed much of the vehicle's kinetic energy at a reduced effective speed.

Taken together, the negative SHAP values should not be interpreted as wet roads with speeding being inherently safer. Rather, this pattern warrants careful interpretation and should be treated with caution. Given the heterogeneity embedded within the speeding violation variable, the potential influence of reporting and coding conventions, the absence of a clear divergence between speeding and non-speeding groups on wet surfaces, and the relatively narrow SHAP value ranges observed, this interaction effect requires further investigation with more granular speed data before any definitive behavioral or policy conclusions can be drawn.

### 5.4.3 Speeding and Lighting Conditions

The SHAP interaction plot showing the combined effect of speeding and lighting conditions on tree-involved crash severity is depicted in Figure 9.

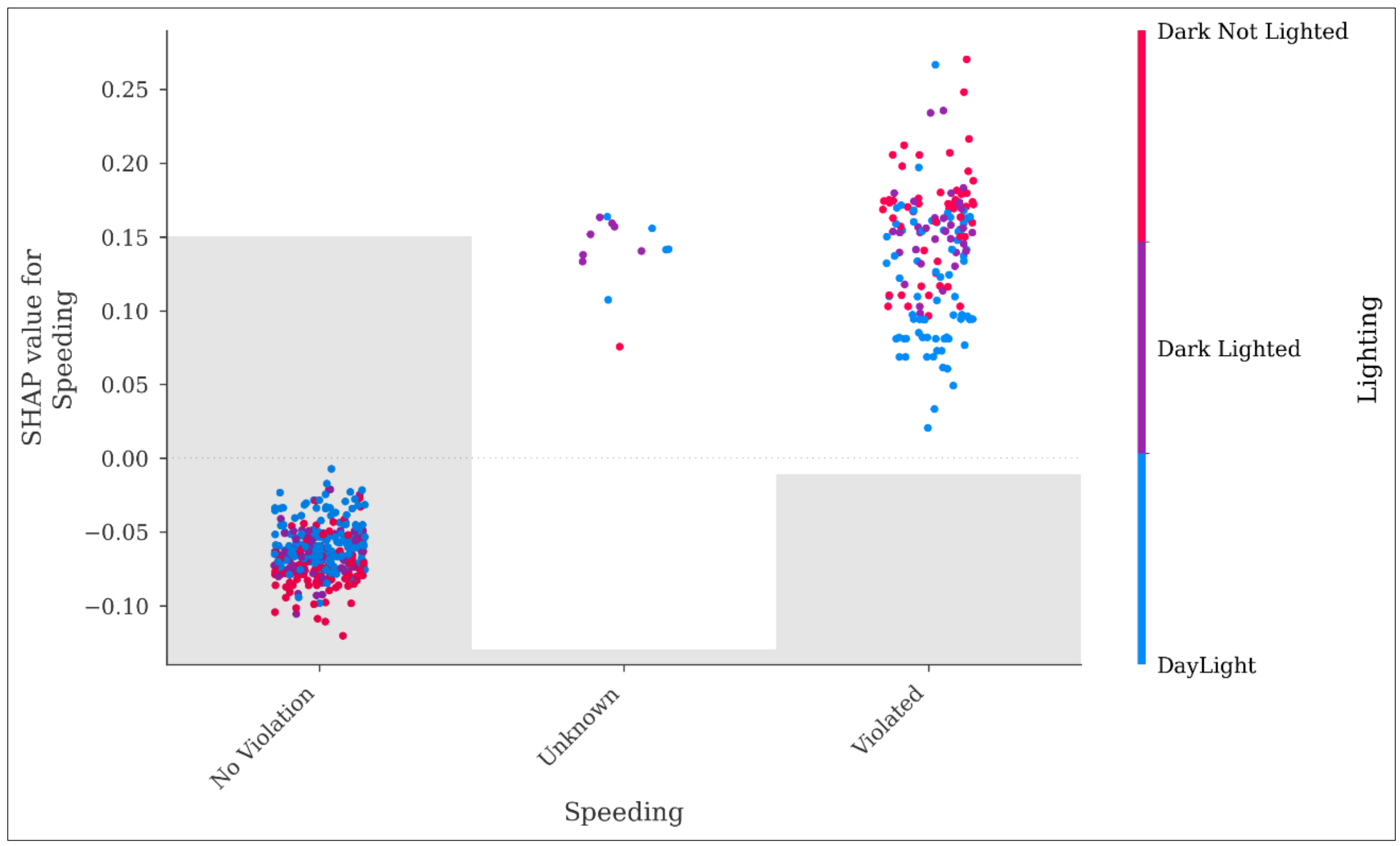

**Figure 9. SHAP Interaction Plot for Speeding and Lighting Conditions**

This SHAP interaction plot reveals how the effect of speeding violations on crash severity is moderated by lighting conditions in tree-involved collisions. When no speeding violation is present, lighting conditions show minimal differentiation, with daylight, dark-lighted, and dark-not-lighted crashes clustering together around negative SHAP values (-0.05 to -0.10), indicating that appropriate speed management provides protective benefits regardless of visibility conditions. However, when speeding violations occur, a dramatic interaction emerges. Speeding in daylight (shown in blue) produces moderate positive SHAP values (+0.02 to +0.10), while speeding in dark conditions, particularly in dark-not-lighted environments (shown in red), generates substantially higher positive SHAP values (+0.10 to +0.27), indicating a substantially increased crash severity. This pattern demonstrates that the dangers of speeding are greatly amplified in low-visibility conditions, where excessive speed combines with reduced sight distance and limited hazard perception, creating a compounding effect. Drivers traveling too fast in darkness have insufficient time to detect roadway departures, recognize obstacles like trees, or brake effectively before impact, resulting in higher-energy collisions and more severe occupant injuries compared to speeding violations that occur during daylight hours when visual feedback allows for some degree of hazard recognition and response. Similar patterns have been documented in previous research, which consistently shows that speeding-related crash severity is markedly elevated under dark or low-visibility conditions [75,76].

**5.4.4 Vehicle Age and Restraint Usage**

The SHAP interaction plot showing the combined effect of vehicle age and restraint usage on tree-involved crash severity is depicted in Figure 10.

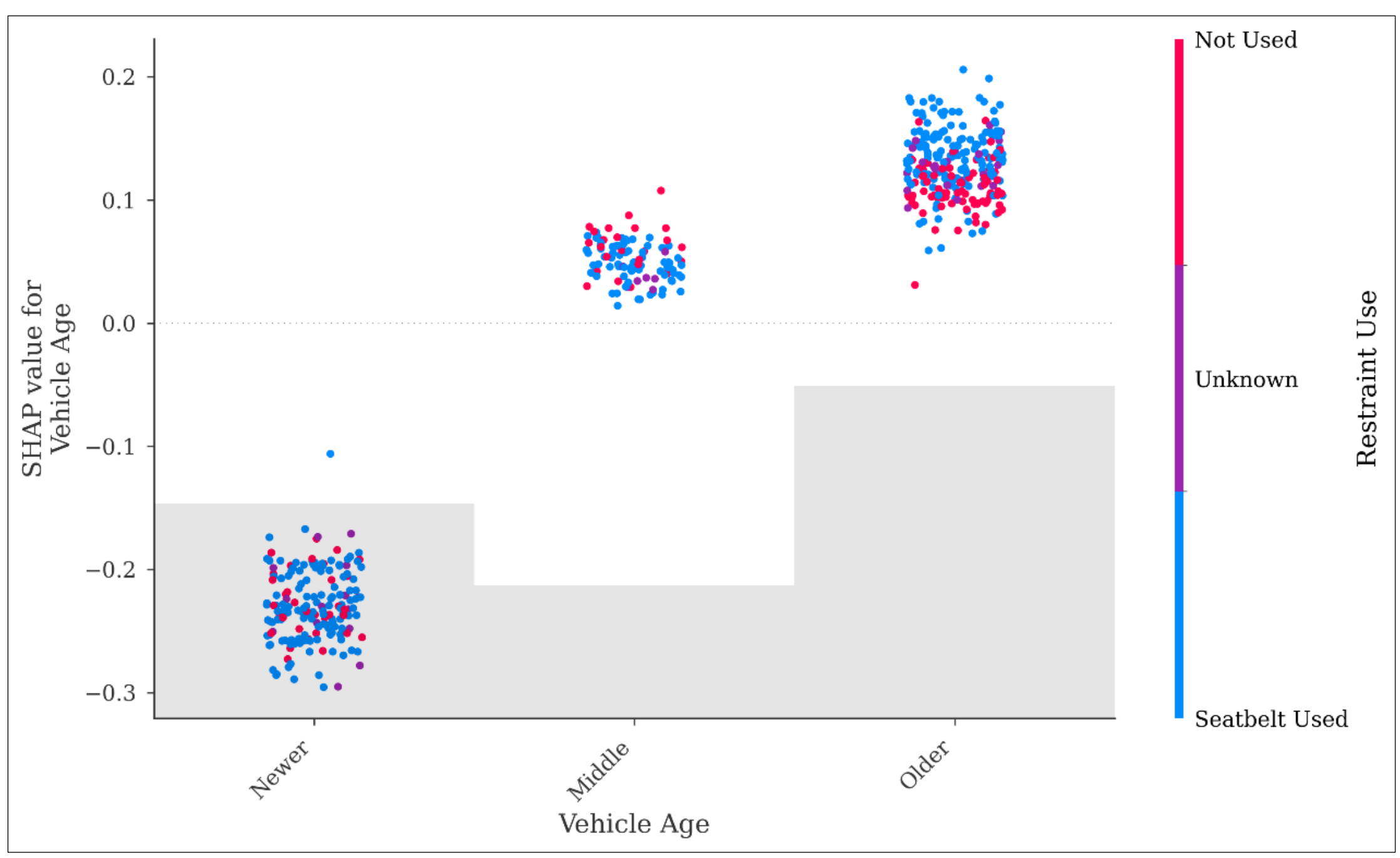

**Figure 10. SHAP Interaction Plot for Vehicle Age and Restraint Usage**

This SHAP interaction plot demonstrates how the effect of vehicle age on crash severity is strongly moderated by restraint use in tree-involved collisions. For newer vehicles, restraint use shows a clear protective effect, with seatbelt-wearing occupants (shown in blue) clustered around negative SHAP values (-0.17 to -0.30) indicating reduced severity, while non-restrained occupants (shown in red) show slightly negative to near-zero values (-0.10 to -0.18), suggesting that modern vehicle safety features provide some baseline protection but are substantially less effective without proper restraint use. As vehicle age increases, the interaction intensifies dramatically with middle-aged and older categories. Older vehicles with unrestrained occupants exhibit strongly positive SHAP values (+0.08 to +0.20), indicating the highest crash severity outcomes, while even restrained occupants in older vehicles show elevated positive SHAP values (+0.06 to +0.18), though still somewhat lower than those of unrestrained occupants. This pattern reveals that vehicle age and restraint use operate synergistically - the safety deficiencies of older vehicles (e.g., reduced structural integrity, less advanced airbag systems, weaker crumple zones) are most catastrophic when combined with non-use of seatbelts, creating a compounding vulnerability where occupants face both inadequate vehicle protection and lack of personal restraint during high-energy tree impacts.

**6. CONCLUSION**

This study examined crash contributing factors in tree-involved motor vehicle collisions across the United States using a hybrid modeling approach combining ML, SHAP interpretability analysis, and a binary logistic regression model. Using nationally representative CRSS data spanning 2020–2023, the analysis systematically identified both shared predictors and context-specific mechanisms underlying injury severity in tree-involved crashes. This research advances crash severity analysis by integrating ML's pattern recognition with statistical modeling's inferential transparency.

The analysis revealed that human behavioral factors emerged as the most dominant predictors of tree-involved crash severity. Restraint use was the single most influential variable, with unrestrained occupants facing dramatically elevated risk of fatal or serious injury due to the sequential and high-energy nature of run-off-road tree crashes. Driver impairment, whether alcohol or drug-related or physical in nature, and speeding were similarly strong contributors, reflecting how these behaviors compound crash risk by degrading vehicle control, reducing reaction time, and increasing impact energy at the point of tree contact. Alcohol involvement further amplified severity risk through its well-documented effects on judgment, visual processing, and speed regulation. Lighting conditions also played a meaningful role, with dark environments limiting drivers' ability to detect roadway hazards early enough to take corrective action, resulting in higher-speed departures and more severe impacts.

Vehicle-related and environmental factors provided additional explanatory value, though with comparatively more modest effect sizes. Older vehicles were associated with greater crash severity, reflecting the cumulative safety benefits of modern structural design, crashworthiness standards, and active safety technologies present in newer models. Airbag deployment, while counterintuitively linked to more severe outcomes, functioned primarily as a proxy for high-energy impacts rather than a causal contributor to injury. Adverse road surface conditions, particularly wet pavement, reduced tire-pavement friction and impaired drivers' ability to brake or maneuver before striking a tree. The location of crashes at non-intersection segments, where trees are most found along the roadside, was also associated with higher severity, though this pattern reflects the spatial distribution of roadside trees rather than an inherent safety characteristic of road geometry. Collectively, these findings underscore the multifaceted nature of tree-involved crash severity and highlight the critical importance of behavioral interventions, fleet modernization, and targeted roadside tree management in reducing the frequency and consequences of these crashes.

The binary logistic regression and SHAP-based analyses yield largely consistent findings, collectively reinforcing the dominant role of behavioral and vehicle-related factors in determining crash severity in tree-involved collisions. Restraint non-use emerged as the strongest predictor across both approaches, with unrestrained occupants nearly three times more likely to sustain fatal or incapacitating injuries, corroborated by the highest positive SHAP values in the ensemble model. Speeding violations, alcohol and drug impairment, and older vehicle age were similarly significant in both frameworks, confirming that higher impact energy, reduced driver control, and diminished vehicle crashworthiness consistently elevate injury severity. Airbag deployment showed elevated odds ratios and positive SHAP values across all deployment categories, though both models converge on the interpretation that deployment functions as a marker of crash energy rather than a causal contributor to injury. While lighting conditions, road surface characteristics, and intersection location demonstrated moderate and broadly consistent effects across both approaches, driver age effects were comparatively modest and more context-dependent. Overall, the two methods mutually validate one another, with the logistic regression providing interpretable effect sizes and the SHAP analysis offering additional distributional insights that extend beyond what regression coefficients alone can convey.

The SHAP interaction plots collectively reveal that crash severity in tree-involved collisions is governed not by individual risk factors in isolation, but by the compounding effects of their co-occurrence. The road surface condition and speeding interaction demonstrates that while speeding on dry surfaces carries moderate additional severity risk, the combination of wet road conditions with no speeding violation yields the most protective outcome, as appropriate speed reduction on low-friction surfaces allows drivers to maintain vehicle control and reduce impact energy. Conversely, speeding on wet surfaces eliminates this protective margin by preventing effective braking and steering during roadway departures. The speeding and lighting condition interaction further illustrates this multiplicative dynamic, where appropriate speed management produces protective SHAP values regardless of visibility conditions, but speeding in dark-not-lighted environments generates the most extreme positive SHAP values observed across all interaction plots, reflecting the catastrophic convergence of reduced sight distance, limited hazard perception, and excessive vehicle speed that leaves drivers with virtually no opportunity to detect or respond to an impending tree collision.

The remaining two interactions introduce vehicle-level and behavioral dimensions that further deepen this understanding. The lighting condition and vehicle age interaction suggests that drivers of newer vehicles may exhibit risk compensation tendencies in dark conditions, partially offsetting the inherent safety advantages of modern vehicle technology, as the false sense of security afforded by advanced safety features may encourage higher speeds and reduced caution in low-visibility environments. The vehicle age and restraint use interaction reveals perhaps the most consequential compounding vulnerability, where the structural deficiencies of older vehicles - reduced crashworthiness, less advanced airbag systems, and weaker crumple zones - are most catastrophic when combined with restraint non-use, producing the highest positive SHAP values observed across all interaction analyses. Even restrained occupants in older vehicles face elevated severity risk, though meaningfully lower than their unrestrained counterparts, while newer vehicles provide meaningful baseline protection that is nonetheless substantially diminished without proper seatbelt use. Taken together, these interactions underscore that the most severe tree-involved crashes arise from the simultaneous presence of multiple compounding risk factors, and that interventions targeting any single factor in isolation may yield limited benefits unless the broader constellation of co-occurring conditions is also addressed.

### 6.1 Safety Recommendations

The findings highlight several actionable intervention points across roadway design, enforcement, vehicle technology, and road-user behavior. Given that tree-involved crashes are disproportionately severe and strongly shaped by preventable factors such as restraint non-use and alcohol involvement, a multi-layered safety strategy is essential.

Given that restraint non-use is one of the most influential predictors of KA outcomes in tree-involved crashes, jurisdictions should implement high-visibility nighttime seatbelt enforcement, particularly along tree-lined corridors where run-off-road crashes cluster. Combining seatbelt enforcement with sobriety checkpoints amplifies the impact by concurrently addressing alcohol-related impairment, a common co-factor in severe tree impacts. Public messaging campaigns should explicitly illustrate the fatal biomechanical consequences of tree impacts when restraints are not worn, using crash-reconstruction imagery and narrative case examples to increase perceived risk.

Since tree-involved crashes typically stem from roadway departures, infrastructure countermeasures should prioritize reducing the likelihood and severity of run-off-road crashes. Recommended treatments include shoulder widening, shoulder and centerline rumble strips, and high-friction surface treatments, which have been shown to improve vehicle control and reduce roadway departures in high-risk locations [77,78]. Clear zone management should focus on selectively removing high-risk trees or installing energy-absorbing barriers and guardrails where removal is not feasible, ensuring that vehicles departing the roadway have a more forgiving roadside environment [79].

Older vehicles demonstrated substantially greater severity risk due to inferior structural integrity and outdated safety systems. These risks compound dramatically when restraints are not used, as shown in the SHAP interaction between vehicle age and restraint non-use. Policymakers should consider expanding vehicle retirement incentives, low-income safety upgrade programs, and enhanced inspection protocols prioritizing seatbelt integrity, airbag functionality, and electronic stability control (ESC) availability. Partnerships with insurance providers could accelerate the transition away from high-risk aging vehicles by offering premium discounts tied to modern vehicle safety performance.

Speed management and visibility countermeasures are also critical given the strong associations observed between speeding violations, dark not lighted conditions, and increased crash severity in tree-involved events. Speeding substantially reduces the available reaction time for drivers to correct roadway departures, and the SHAP interaction results from this study demonstrate that these risks are amplified under dark-not-lighted conditions, where limited sight distance diminishes hazard recognition. Targeted nighttime speed enforcement, automated speed monitoring, and context-sensitive speed limits can help address these risks [80]. Improving nighttime visibility through installations of LED-based roadway lighting, part-time/adaptive lighting systems, and enhanced retroreflective delineation can further mitigate the elevated severity associated with dark environments.

Communication strategies should emphasize the biomechanical severity of tree impacts, which function as collisions with non-deformable fixed objects and generate high deceleration forces even at relatively modest travel speeds. Risk-communication tools that employ data-driven visualizations, such as simulation-based crash animations, simplified biomechanical diagrams, and SHAP-derived explanations of key contributing factors, can enhance public understanding of how restraint non-use and older vehicle structures substantially elevate the likelihood of fatal injury. Given their disproportionate involvement in single-vehicle roadway-departure crashes, young drivers should be the primary audience for these campaigns, with messaging tailored to address the specific behavioral patterns and risk perceptions prevalent within this demographic group.

### 6.2 Study Limitations

This study has several limitations that should be considered when interpreting the findings. First, the analysis relies exclusively on police-reported crashes from the CRSS database for the period 2020–2023. Although CRSS is designed to be nationally representative, it is still a sample of all crashes and is subject to under-reporting and reporting biases, particularly for lower-severity events. The constructed analysis dataset is limited by the variables available in CRSS. Important roadside and tree-specific attributes, such as tree species, diameter, height, setback distance from the travel lane, clustering of trees, and presence and condition of guardrails, are not explicitly captured and therefore cannot be incorporated into the modeling framework. In addition, the CRSS data contain a substantial number of "unknown" or "not reported" entries for key variables such as impairment, speeding, restraint use, and some infrastructure characteristics. In this study, these categories were retained and modeled as separate levels to preserve sample size, but any systematic patterns in missingness or misclassification could bias coefficient estimates and SHAP-based importance measures.

## REFERENCES

[1] Federal Highway Administration (FHWA), Tree Crashes, U.S. Department of Transportation, 2021. https://highways.dot.gov/sites/fhwa.dot.gov/files/FHWA-SA-21-022_Tree_Crashes.pdf.

[2] K. Bucsuházy, R. Zůvala, V. Valentová, J. Ambros, Factors related to severe single-vehicle tree crashes: In-depth crash study, PLOS ONE 17 (2022) e0248171. https://doi.org/10.1371/journal.pone.0248171.

[3] F.A. Pintar, N. Yoganandan, D.J. Maiman, Injury mechanisms and severity in narrow offset frontal impacts, Ann. Adv. Automot. Med. Assoc. Adv. Automot. Med. Annu. Sci. Conf. 52 (2008) 185–189.

[4] Austroad, Guide to Road Design Part 6: Roadside Design, Safety and Barriers, Austroads, 2024. https://austroads.gov.au/publications/road-design/agrd06.

[5] F.A. Pintar, D.J. Maiman, N. Yoganandan, Injury Patterns in Side Pole Crashes, Annu. Proc. Assoc. Adv. Automot. Med. 51 (2007) 419–433. https://pmc.ncbi.nlm.nih.gov/articles/PMC3217499/.

[6] D.S. Turner, E.R. Mansfield, Urban Trees and Roadside Safety, J. Transp. Eng. 116 (1990) 90–104. https://doi.org/10.1061/(ASCE)0733-947X(1990)116:1(90).

[7] J.M. Holdridge, V.N. Shankar, G.F. Ulfarsson, The crash severity impacts of fixed roadside objects, J. Safety Res. 36 (2005) 139–147. https://doi.org/10.1016/j.jsr.2004.12.005.

[8] T. Yamamoto, V.N. Shankar, Bivariate ordered-response probit model of driver's and passenger's injury severities in collisions with fixed objects, Accid. Anal. Prev. 36 (2004) 869–876. https://doi.org/10.1016/j.aap.2003.09.002.

[9] A. Daniello, H.C. Gabler, Fatality risk in motorcycle collisions with roadside objects in the United States, Accid. Anal. Prev. 43 (2011) 1167–1170. https://doi.org/10.1016/j.aap.2010.12.027.

[10] M.R. Bambach, R.H. Grzebieta, J. Olivier, A.S. McIntosh, Fatality Risk for Motorcyclists in Fixed Object Collisions, J. Transp. Saf. Secur. 3 (2011) 222–235. https://doi.org/10.1080/19439962.2011.587940.

[11] R. Shrestha, L. Ventura, N. Venkataraman, V. Shankar, An error components mixed logit with heterogeneity in means and variance for fixed object occupant severity outcomes, Anal. Methods Accid. Res. 42 (2024) 100330. https://doi.org/10.1016/j.amar.2024.100330.

[12] Md.M. Rahman, S. Hernandez, R.M. Radwan Albatayneh, Assessing the impact of COVID-19 on driver injury severities in fixed-object passenger car crashes: Insights from temporal and partially constrained modeling analysis, Anal. Methods Accid. Res. 47 (2025) 100397. https://doi.org/10.1016/j.amar.2025.100397.

[13] V. Bendigeri, Analysis of factors contributing to roadside tree crashes in South Carolina, Theses (2009). https://open.clemson.edu/all_theses/711.

[14] S. Das, B. Storey, T.H. Shimu, S. Mitra, M. Theel, B. Maraghehpour, Severity analysis of tree and utility pole crashes: Applying fast and frugal heuristics, IATSS Res. 44 (2020) 85–93. https://doi.org/10.1016/j.iatssr.2019.08.001.

[15] K.L. Wolf, N. Bratton, Urban Trees and Traffic Safety: Considering U.S. Roadside Policy and Crash Data, Arboric. Urban For. AUF 32 (2006) 170. https://doi.org/10.48044/jauf.2006.023.

[16] W.E. Marshall, Y. Golombek, N. Coppola, B. Janson, The Unresolved Relationship between Street Trees and Road Safety, Mountain-Plains Consortium, Fargo, ND, 2019. https://trid.trb.org/View/1650959.

[17] J. Van Treese Ii, A. Koeser, G. Fitzpatrick, M. Olexa, E. Allen, Frequency and Severity of Tree and Other Fixed Object Crashes in Florida, 2006—2013, Arboric. Urban For. 45 (2019). https://doi.org/10.48044/jauf.2019.006.

[18] M.H. Ray, Impact conditions in side-impact collisions with fixed roadside objects, Accid. Anal. Prev. 31 (1999) 21–30. https://doi.org/10.1016/S0001-4575(98)00041-4.

[19] M.H. Ray, K. Hiranmayee, Evaluating Human Risk in Side Impact Collisions with Roadside Objects, Transp. Res. Rec. 1720 (2000) 67–71. https://doi.org/10.3141/1720-08.

[20] C.L. Naing, J. Hill, R. Thomson, H. Fagerlind, M. Kelkka, C. Klootwijk, G. Dupre, O. Bisson, Single-vehicle collisions in Europe: analysis using real-world and crash-test data, Int. J. Crashworthiness 13 (2008) 219–229. https://doi.org/10.1080/13588260701788583.
[21] C.D. Fitzpatrick, The Effect of Roadside Elements on Driver Behavior and Run-Off-the-Road Crash Severity, Doctoral Dissertation, University of Massachusetts Amherst, 2013. https://scholarworks.umass.edu/dissertations/AAI3603093.
[22] J.W. Van Treese II, A.K. Koeser, G.E. Fitzpatrick, M.T. Olexa, E.J. Allen, A review of the impact of roadway vegetation on drivers' health and well-being and the risks associated with single-vehicle crashes, Arboric. J. 39 (2017) 179–193. https://doi.org/10.1080/03071375.2017.1374591.
[23] G. Cheng, R. Cheng, Y. Pei, L. Xu, W. Qi, Severity assessment of accidents involving roadside trees based on occupant injury analysis, PLOS ONE 15 (2020) e0231030. https://doi.org/10.1371/journal.pone.0231030.
[24] N.V. Malyshkina, F.L. Mannering, Markov switching multinomial logit model: An application to accident-injury severities, Accid. Anal. Prev. 41 (2009) 829–838. https://doi.org/10.1016/j.aap.2009.04.006.
[25] K.M. Kockelman, Y.-J. Kweon, Driver injury severity: an application of ordered probit models, Accid. Anal. Prev. 34 (2002) 313–321. https://doi.org/10.1016/S0001-4575(01)00028-8.
[26] F. Wei, G. Lovegrove, An empirical tool to evaluate the safety of cyclists: Community based, macro-level collision prediction models using negative binomial regression, Accid. Anal. Prev. 61 (2013) 129–137. https://doi.org/10.1016/j.aap.2012.05.018.
[27] A.S. Al-Ghamdi, Using logistic regression to estimate the influence of accident factors on accident severity, Accid. Anal. Prev. 34 (2002) 729–741. https://doi.org/10.1016/S0001-4575(01)00073-2.
[28] D.W. Kononen, C.A.C. Flannagan, S.C. Wang, Identification and validation of a logistic regression model for predicting serious injuries associated with motor vehicle crashes, Accid. Anal. Prev. 43 (2011) 112–122. https://doi.org/10.1016/j.aap.2010.07.018.
[29] Y. Li, R. Gu, J. Lee, M. Yang, Q. Chen, Y. Zhang, The dynamic tradeoff between safety and efficiency in discretionary lane-changing behavior: A random parameters logit approach with heterogeneity in means and variances, Accid. Anal. Prev. 153 (2021) 106036. https://doi.org/10.1016/j.aap.2021.106036.
[30] S. Zhang, N.N. Sze, Real-time conflict risk at signalized intersection using drone video: A random parameters logit model with heterogeneity in means and variances, Accid. Anal. Prev. 207 (2024) 107739. https://doi.org/10.1016/j.aap.2024.107739.
[31] A. Hossain, X. Sun, S. Das, M. Jafari, A. Rahman, Investigating pedestrian-vehicle crashes on interstate highways: Applying random parameter binary logit model with heterogeneity in means, Accid. Anal. Prev. 199 (2024) 107503. https://doi.org/10.1016/j.aap.2024.107503.
[32] Y. Ali, F. Hussain, M.M. Haque, Advances, challenges, and future research needs in machine learning-based crash prediction models: A systematic review, Accid. Anal. Prev. 194 (2024) 107378. https://doi.org/10.1016/j.aap.2023.107378.
[33] M. Yan, Y. Shen, Traffic Accident Severity Prediction Based on Random Forest, Sustainability 14 (2022). https://doi.org/10.3390/su14031729.
[34] C. Chen, G. Zhang, Z. Qian, R.A. Tarefder, Z. Tian, Investigating driver injury severity patterns in rollover crashes using support vector machine models, Accid. Anal. Prev. 90 (2016) 128–139. https://doi.org/10.1016/j.aap.2016.02.011.
[35] M. Zheng, T. Li, R. Zhu, J. Chen, Z. Ma, M. Tang, Z. Cui, Z. Wang, Traffic Accident's Severity Prediction: A Deep-Learning Approach-Based CNN Network, IEEE Access 7 (2019) 39897–39910. https://doi.org/10.1109/ACCESS.2019.2903319.
[36] J. Niyogisubizo, L. Liao, Q. Sun, E. Nziyumva, Y. Wang, L. Luo, S. Lai, E. Murwanashyaka, Predicting Crash Injury Severity in Smart Cities: a Novel Computational Approach with Wide and Deep Learning Model, Int. J. Intell. Transp. Syst. Res. 21 (2023) 240–258. https://doi.org/10.1007/s13177-023-00351-7.

[37] G. Antariksa, R. Tamakloe, J. Liu, S. Das, Automated and Explainable Artificial Intelligence to Enhance Prediction of Pedestrian Injury Severity, IEEE Trans. Intell. Transp. Syst. 26 (2025) 5568–5584. https://doi.org/10.1109/TITS.2025.3526217.

[38] Z. Wang, H. Guo, C. Zhang, Z. Hu, F. Zhou, Z. Sun, R. Sherony, S. Bao, Investigating pedestrian crash injury patterns: A comparative study of children and non-children, Accid. Anal. Prev. 222 (2025) 108223. https://doi.org/10.1016/j.aap.2025.108223.

[39] M. Feng, J. Zhao, C. Hou, C. Nie, J. Hou, Investigating the safety influence path of right-turn configurations on vehicle–pedestrian conflict risk at signalized intersections, Accid. Anal. Prev. 211 (2025) 107910. https://doi.org/10.1016/j.aap.2024.107910.

[40] A. Agheli, K. Aghabayk, How does distraction affect cyclists' severe crashes? A hybrid CatBoost-SHAP and random parameters binary logit approach, Accid. Anal. Prev. 211 (2025) 107896. https://doi.org/10.1016/j.aap.2024.107896.

[41] A. Goswamy, M. Abdel-Aty, Z. Islam, Factors affecting injury severity at pedestrian crossing locations with Rectangular RAPID Flashing Beacons (RRFB) using XGBoost and random parameters discrete outcome models, Accid. Anal. Prev. 181 (2023) 106937. https://doi.org/10.1016/j.aap.2022.106937.

[42] Z. Sun, D. Wang, X. Gu, M. Abdel-Aty, Y. Xing, J. Wang, H. Lu, Y. Chen, A hybrid approach of random forest and random parameters logit model of injury severity modeling of vulnerable road users involved crashes, Accid. Anal. Prev. 192 (2023) 107235. https://doi.org/10.1016/j.aap.2023.107235.

[43] A. Scarano, M. Rella Riccardi, F. Mauriello, C. D'Agostino, N. Pasquino, A. Montella, Injury severity prediction of cyclist crashes using random forests and random parameters logit models, Accid. Anal. Prev. 192 (2023) 107275. https://doi.org/10.1016/j.aap.2023.107275.

[44] S. Azmeri Khan, S. Yasmin, M. Mazharul Haque, Effects of design consistency measures and roadside hazard types on run-off-road crash severity: Application of random parameters hierarchical ordered probit model, Anal. Methods Accid. Res. 40 (2023) 100300. https://doi.org/10.1016/j.amar.2023.100300.

[45] M. Sadeghi, K. Aghabayk, M. Quddus, A hybrid Machine learning and statistical modeling approach for analyzing the crash severity of mobility scooter users considering temporal instability, Accid. Anal. Prev. 206 (2024) 107696. https://doi.org/10.1016/j.aap.2024.107696.

[46] A. Hossain, X. Sun, S. Das, M. Jafari, J. Codjoe, Investigating older driver crashes on high-speed roadway segments: a hybrid approach with extreme gradient boosting and random parameter model, Transp. Transp. Sci. 22 (2024) 1–35. https://doi.org/10.1080/23249935.2024.2362362.

[47] A.V. Dorogush, V. Ershov, A. Gulin, CatBoost: gradient boosting with categorical features support, (2018). https://doi.org/10.48550/arXiv.1810.11363.

[48] L. Prokhorenkova, G. Gusev, A. Vorobev, A.V. Dorogush, A. Gulin, CatBoost: unbiased boosting with categorical features, in: Adv. Neural Inf. Process. Syst., Curran Associates, Inc., 2018. https://proceedings.neurips.cc/paper_files/paper/2018/hash/14491b756b3a51daac41c24863285549-Abstract.html.

[49] S.M. Lundberg, S.-I. Lee, A unified approach to interpreting model predictions, in: Proc. 31st Int. Conf. Neural Inf. Process. Syst., Curran Associates Inc., Red Hook, NY, USA, 2017: pp. 4768–4777.

[50] S.M. Lundberg, G. Erion, H. Chen, A. DeGrave, J.M. Prutkin, B. Nair, R. Katz, J. Himmelfarb, N. Bansal, S.-I. Lee, From local explanations to global understanding with explainable AI for trees, Nat. Mach. Intell. 2 (2020) 56–67. https://doi.org/10.1038/s42256-019-0138-9.

[51] S. Washington, M.G. Karlaftis, F. Mannering, P. Anastasopoulos, Statistical and Econometric Methods for Transportation Data Analysis, 3rd ed., Chapman and Hall/CRC, New York, 2020. https://doi.org/10.1201/9780429244018.

[52] D.W. Hosmer, S. Lemeshow, R.X. Sturdivant, Applied Logistic Regression, 3. Aufl, Wiley, Hoboken, N.J, 2013.

[53] I. Mohamad, Quantifying the life-saving impact of seatbelt usage: A random forest analysis of unobserved heterogeneity and latent risk factors in vehicular fatalities, Multimodal Transp. 4 (2025) 100221. https://doi.org/10.1016/j.multra.2025.100221.
[54] N. Fouda Mbarga, A.-R. Abubakari, L.N. Aminde, A.R. Morgan, Seatbelt use and risk of major injuries sustained by vehicle occupants during motor-vehicle crashes: a systematic review and meta-analysis of cohort studies, BMC Public Health 18 (2018) 1413. https://doi.org/10.1186/s12889-018-6280-1.
[55] V. Sarwahi, A.M. Atlas, J. Galina, A. Satin, T.J.I. Dowling, S. Hasan, T.D. Amaral, Y. Lo, N. Christopherson, J. Prince, Seatbelts Save Lives, and Spines, in Motor Vehicle Accidents: A Review of the National Trauma Data Bank in the Pediatric Population, Spine 46 (2021) 1637. https://doi.org/10.1097/BRS.0000000000004072.
[56] C. Lee, X. Li, Predicting Driver Injury Severity in Single-Vehicle and Two-Vehicle Crashes with Boosted Regression Trees, Transp. Res. Rec. 2514 (2015) 138–148. https://doi.org/10.3141/2514-15.
[57] S. Islam, A.B. Hossain, T.E. Barnett, Comprehensive Injury Severity Analysis of SUV and Pickup Truck Rollover Crashes: Alabama Case Study, Transp. Res. Rec. 2601 (2016) 1–9. https://doi.org/10.3141/2601-01.
[58] F.L. Mannering, V. Shankar, C.R. Bhat, Unobserved heterogeneity and the statistical analysis of highway accident data, Anal. Methods Accid. Res. 11 (2016) 1–16. https://doi.org/10.1016/j.amar.2016.04.001.
[59] G. Fountas, P.Ch. Anastasopoulos, F.L. Mannering, Analysis of vehicle accident-injury severities: A comparison of segment- versus accident-based latent class ordered probit models with class-probability functions, Anal. Methods Accid. Res. 18 (2018) 15–32. https://doi.org/10.1016/j.amar.2018.03.003.
[60] D. Shannon, F. Murphy, M. Mullins, L. Rizzi, Exploring the role of delta-V in influencing occupant injury severities – A mediation analysis approach to motor vehicle collisions, Accid. Anal. Prev. 142 (2020) 105577. https://doi.org/10.1016/j.aap.2020.105577.
[61] R. Arvin, A.J. Khattak, Driving impairments and duration of distractions: Assessing crash risk by harnessing microscopic naturalistic driving data, Accid. Anal. Prev. 146 (2020) 105733. https://doi.org/10.1016/j.aap.2020.105733.
[62] S.M. Simmons, M. Donoghue, S. Erdelyi, H. Chan, C. Vaillancourt, P. Atkinson, F. Besserer, D.B. Clarke, P. Davis, R. Daoust, M. Émond, J. Eppler, J.S. Lee, A. MacPherson, K. Magee, E. Mercier, R. Ohle, M. Parsons, J. Rao, B.H. Rowe, J. Taylor, I. Wishart, J.R. Brubacher, Influence of cannabis and alcohol on motor vehicle injury severity in Canadian trauma centres: a prospective study, Inj. Prev. (2025). https://doi.org/10.1136/ip-2025-045642.
[63] M. Islam, P. Hosseini, A. Kakhani, M. Jalayer, D. Patel, Unveiling the risks of speeding behavior by investigating the dynamics of driver injury severity through advanced analytics, Sci. Rep. 14 (2024) 22431. https://doi.org/10.1038/s41598-024-73134-z.
[64] M. Islam, A. Mahmud, Unveiling the speeding behavior: Assessing the speeding risks and driver injury severities in single-heavy truck crashes, Saf. Sci. 187 (2025) 106861. https://doi.org/10.1016/j.ssci.2025.106861.
[65] Y. Chen, Y. Li, M. King, Q. Shi, C. Wang, P. Li, Identification methods of key contributing factors in crashes with high numbers of fatalities and injuries in China, Traffic Inj. Prev. 17 (2016) 878–883. https://doi.org/10.1080/15389588.2016.1174774.
[66] A. Hossain, X. Sun, S. Islam, S. Alam, Md. Mahmud Hossain, Identifying roadway departure crash patterns on rural two-lane highways under different lighting conditions: Association knowledge using data mining approach, J. Safety Res. 85 (2023) 52–65. https://doi.org/10.1016/j.jsr.2023.01.006.
[67] R. Chakraborty, J. Liu, A.G. Tusti, M.S. Mimi, S. Das, Impact of lighting conditions on nighttime crash severity among older and elderly drivers, J. Transp. Saf. Secur. 17 (2025) 1377–1417. https://doi.org/10.1080/19439962.2025.2529833.

[68] A. Jafari Anarkooli, M. Hadji Hosseinlou, Analysis of the injury severity of crashes by considering different lighting conditions on two-lane rural roads, J. Safety Res. 56 (2016) 57–65. https://doi.org/10.1016/j.jsr.2015.12.003.

[69] B. Roudsari, R. Kaufman, R. Nirula, Comparison of mid-block and intersection-related left turn collisions, Traffic Inj. Prev. 8 (2007) 393–397. https://doi.org/10.1080/15389580701603227.

[70] M. Asgarzadeh, S. Verma, R.A. Mekary, T.K. Courtney, D.C. Christiani, The role of intersection and street design on severity of bicycle-motor vehicle crashes, Inj. Prev. J. Int. Soc. Child Adolesc. Inj. Prev. 23 (2017) 179–185. https://doi.org/10.1136/injuryprev-2016-042045.

[71] Md.M. Haque, H.C. Chin, H. Huang, Modeling fault among motorcyclists involved in crashes, Accid. Anal. Prev. 41 (2009) 327–335. https://doi.org/10.1016/j.aap.2008.12.010.

[72] Y. Li, J. Huang, Safety Impact of Pavement Conditions, Transp. Res. Rec. 2455 (2014) 77–88. https://doi.org/10.3141/2455-09.

[73] A.P. Afghari, M.M. Haque, S. Washington, T. Smyth, Bayesian Latent Class Safety Performance Function for Identifying Motor Vehicle Crash Black Spots, Transp. Res. Rec. 2601 (2016) 90–98. https://doi.org/10.3141/2601-11.

[74] S.R. Vertlib, S. Rosenzweig, O.D. Rubin, A. Steren, Are car safety systems associated with more speeding violations? Evidence from police records in Israel, PLOS ONE 18 (2023) e0286622. https://doi.org/10.1371/journal.pone.0286622.

[75] J. Liu, J. Li, K. Wang, J. Zhao, H. Cong, P. He, Exploring factors affecting the severity of night-time vehicle accidents under low illumination conditions, Adv. Mech. Eng. 11 (2019) 1687814019840940. https://doi.org/10.1177/1687814019840940.

[76] K. Zhang, M. Hassan, Crash severity analysis of nighttime and daytime highway work zone crashes, PLOS ONE 14 (2019) e0221128. https://doi.org/10.1371/journal.pone.0221128.

[77] C. Lyon, B. Persaud, D. Merritt, J. Cheung, Empirical Bayes Before-After Study to Develop Crash Modification Factors and Functions for High Friction Surface Treatments on Curves and Ramps, Transp. Res. Rec. 2674 (2020) 505–514. https://doi.org/10.1177/0361198120957327.

[78] G. Cheng, R. Cheng, Y. Pei, J. Han, Research on Highway Roadside Safety, J. Adv. Transp. 2021 (2021) 6622360. https://doi.org/10.1155/2021/6622360.

[79] F.D.B. de Albuquerque, D.M. Awadalla, Roadside Fixed-Object Collisions, Barrier Performance, and Fatal Injuries in Single-Vehicle, Run-Off-Road Crashes, Safety 6 (2020) 27. https://doi.org/10.3390/safety6020027.

[80] H. Sadeghi-Bazargani, M. Saadati, Speed Management Strategies; A Systematic Review, Bull. Emerg. Trauma 4 (2016) 126–133. https://pmc.ncbi.nlm.nih.gov/articles/PMC4989038/.